# Inference for Sparse Conditional Precision Matrices

Jialei Wang* and Mladen Kolar†

**Abstract**

Given $n$ i.i.d. observations of a random vector $(X, Z)$, where $X$ is a high-dimensional vector and $Z$ is a low-dimensional index variable, we study the problem of estimating the conditional inverse covariance matrix $\Omega(z) = \left(E[(X - E[X \mid Z])(X - E[X \mid Z])^T \mid Z = z]\right)^{-1}$ under the assumption that the set of non-zero elements is small and does not depend on the index variable. We develop a novel procedure that combines the ideas of the local constant smoothing and the group Lasso for estimating the conditional inverse covariance matrix. A proximal iterative smoothing algorithm is used to solve the corresponding convex optimization problems. We prove that our procedure recovers the conditional independence assumptions of the distribution $X \mid Z$ with high probability. This result is established by developing a uniform deviation bound for the high-dimensional conditional covariance matrix from its population counterpart, which may be of independent interest. Furthermore, we develop point-wise confidence intervals for individual elements of the conditional inverse covariance matrix. We perform extensive simulation studies, in which we demonstrate that our proposed procedure outperforms sensible competitors. We illustrate our proposal on a S&P 500 stock price data set.

## 1 Introduction

In recent years, we have witnessed advancement of large-scale data acquisition technology in numerous engineering, scientific and socio-economical domains. Data collected in these domains are complex, noisy and high-dimensional. Graphical modeling of complex dependencies and associations underlying the system that generated the data has become an important tool for practitioners interested in gaining insights into complicated data generating mechanisms. The central quantity of interest in graphical modeling is the inverse covariance matrix $\Omega = \Sigma^{-1}$, since the pattern of non-zero elements in $\Omega$ characterize the edge set in the graphical representation of the system. In particular, each measured variable is represented by a

*Department of Computer Science, The University of Chicago, Chicago, IL 60637, USA; e-mail: jialei@uchicago.edu.
†The University of Chicago Booth School of Business, Chicago, IL 60637, USA; e-mail: mkolar@chicagobooth.edu.

This work is supported in part by an IBM Corporation Faculty Research Fund at the University of Chicago Booth School of Business. This work was completed in part with resources provided by the University of Chicago Research Computing Center.

node and an edge is put between the nodes if the corresponding element in the inverse covariance matrix is non-zero. Unfortunately, the sample covariance matrix $\widehat{\Sigma}$ is a poor estimate of the covariance matrix $\Sigma$ in the high-dimensional regime when the number of variables $p$ is comparable to or much larger than the sample size $n$, as is common in contemporary applications. Furthermore, when the number of variables is larger than the sample size, the sample covariance matrix is not invertible, which further complicates the task of estimating the inverse covariance matrix $\Omega$. Due to its importance, rich literature has emerged on the problem of estimating $\Sigma$ and $\Omega$ in both the statistical and machine learning community (see, for example, Pourahmadi, 2013, for a recent overview).

In many scientific domains it is believed that only a few partial correlations between objects in a system are significant, while others are negligible in comparison. This leads to an assumption that the inverse covariance matrix is sparse, which means that only a few elements are non-zero. There is a vast and rich body of work on the problem of *covariance selection*, which was introduced in Dempster (1972), in recent literature (see, for example, Meinshausen and Bühlmann, 2006, Banerjee et al., 2008, Friedman et al., 2008, Ravikumar et al., 2011). However, most of the current focus and progress is restricted to simple scenarios where the inverse covariance matrix is studied in isolation, without considering effects of other covariates that represent environmental factors. In this paper, we study the problem of *conditional covariance selection* in a high-dimensional setting.

Let $(X, Z)$ be a pair of random variables where $X \in \mathbb{R}^p$ and $Z \in \mathbb{R}$. The problem of conditional covariance selection can be simply described as the estimation of the conditional inverse covariance matrix $\Omega(z) = \left(E[(X - E[X \mid Z])(X - E[X \mid Z])^T \mid Z = z]\right)^{-1}$ with zeros. The random variable $Z$ is called the index variable and the matrix $\Omega(z)$ explains associations between different components of the vector $X$ as function of the index variable $Z$. In this paper, we estimate $\Omega(z)$ using a penalized local Gaussian-likelihood approach, which is suitable in a setting where little is known about the functional relationship between the index variable $Z$ and the associations between components of $X$. As we will show later in the paper, our procedure allows for estimation of $\Omega(z)$ without making strong assumptions on the distribution $(X, Z)$ and only smoothness assumption on $z \mapsto \Omega(z)$. The penalty function used in our approach is based on the $\ell_2$ norm, which encourages the estimated sparsity pattern of $\Omega(z)$ to be the same for all values of the index variable while still allowing for the strength of association to change. Fixing the sparsity pattern of $\Omega(z)$, as a function of $z$, reduces the complexity of the model class and allows us to estimate the conditional inverse covariance matrix more reliably by better utilizing the sample. In addition to developing an estimation procedure that works with minimal assumptions, we also focus on statistical properties of the estimator in the high-dimensional setting, where the number of dimensions $p$ is comparable or even larger than the sample size $n$. The high-dimensional aspect of the problem forces us to carefully analyze statistical properties of the estimator, that would otherwise be apparent in a low-dimensional setting. We prove that our procedure recovers the conditional independence assumptions of the distribution $X \mid Z$ with high probability under suitable technical conditions. The key result that allows us to establish sparsistency of our procedure is a uniform deviation bound for the high-dimensional conditional covariance matrix, which may be of independent interest. Finally, we develop point-wise confidence intervals for individual elements of the conditional inverse covariance matrix.

Our paper contributes to the following literature (i) high-dimensional estimation of in-verse covariance matrices, (ii) non-parametric estimation of covariance matrices, (iii) high-dimensional estimation of varying-coefficient models, and (iv) high dimensional inference. In the classical problem of covariance selection (Dempster, 1972) the inverse covariance matrix does not depend on the index variable, which should be distinguished from our approach. A number of recently developed methods address the problem of covariance selection in high dimensions. These methods are based on maximizing the penalized likelihood (Yuan and Lin, 2007, Rothman et al., 2008, Ravikumar et al., 2011, Friedman et al., 2008, Banerjee et al., 2008, Duchi et al., 2008, Lam and Fan, 2009, Yuan, 2012, Mazumder and Hastie, 2012, Witten et al., 2011, Cai et al., 2012) or pseudo-likelihood (Meinshausen and Bühlmann, 2006, Peng et al., 2009, Rocha et al., 2008, Yuan, 2010, Cai et al., 2011, Sun and Zhang, 2012, Khare et al., 2013). Many interesting problems cannot be cast in the context of covariance selection from independent and identically distributed data. As a result, a number of authors have proposed ways to estimate sparse inverse covariance matrices and corresponding graphs in more challenging settings. Guo et al. (2011), Varoquaux et al. (2010), Honorio and Samaras (2010), Chiquet et al. (2011), Danaher et al. (2014) and Mohan et al. (2014) study estimation of multiple related Gaussian graphical models in a multi-task setting. While our work is related to this body of literature, there are some fundamental differences. Instead of estimating a small, fixed number of inverse covariance matrices that are related in some way, we are estimating a function $z \mapsto \Omega(z)$, which is an infinite dimensional object and requires a different treatment. Closely related is work on estimation of high-dimensional time-varying graphical models in Zhou et al. (2010), Kolar et al. (2010b), Kolar and Xing (2012), Kolar and Xing (2009) and Chen et al. (2013b), where authors focus on estimating $\Omega(z)$ at a fixed time point. In contrast, we estimate the whole mapping $z \mapsto \Omega(z)$ under the assumption that the sparsity pattern does not change with the index variable. Zhou et al. (2010) and Kolar et al. (2010b) allow the sparsity pattern of $\Omega(z)$ to change with the index variable, however, when data are scarce, it is sensible to restrict the model class within which we perform estimation. Recent literature on conditional covariance estimation includes Yin et al. (2010) and Hoff and Niu (2012), however, these authors do not consider ultra-high dimensional setting. Kolar et al. (2010a) uses a pseudolikelihood approach to learn structure of conditional graphical model. Our work is also different from the literature on sparse conditional Gaussian graphical models (see, for example, Yin and Li, 2011, Li et al., 2012, Chen et al., 2013a, Cai et al., 2013, Chun et al., 2013, Zhu et al., 2013, Yin and Li, 2013), where it is assumed that $X \mid Z = z \sim N(\Gamma z, \Sigma)$. In particular, the covariates $Z$ only affect the conditional mean, where in our work the covariance matrix depends on the indexing variable $Z$.

Our work also contributes to the literature on varying-coefficient models (see, for example, Fan and Zhang, 2008, Park et al., 2013, for recent survey articles). Work on variable selection in high-dimensional varying-coefficient models is focused on generalizing the linear regression where the relationship between the response and predictors is linear after conditioning on the index variable (see, for example, Wang et al., 2008, Liu et al., 2013, Tang et al., 2012). In this paper, after conditioning on the index variable the model is multivariate Normal and the standard techniques cannot be directly applied. Furthermore, we develop point-wise confidence intervals for individual parameters in the model, which has not been done in the context of high-dimensional varying coefficient models and requires different techniques to those used in Zhang and Lee (2000). Only recently have methods for constructing confidence intervals for individual parameters in high-dimensional models been proposed (see, for example, Zhang and Zhang, 2011, Javanmard and Montanari, 2013, van de Geer et al., 2013). We construct our point-wise confidence intervals using a similar approach to one described in Jankova and van de Geer (2014) (alternative constructions were given in Liu 2013 and Ren et al. 2013). The basic idea is to construct a de-sparsified estimator, in which the bias introduced by the sparsity inducing penalty is removed, and prove that this estimator is asymptotically Normal. Additional difficulty in our paper is the existence of the bias term that arises due to local smoothing.

In summary, here are the highlights of our paper. Our main contribution is the development of a new procedure for estimation of the conditional inverse covariance matrix under an assumption that the set of non-zero elements of $\Omega(z)$ does not change with the index variable. Our procedure is based on maximizing the penalized local Gaussian likelihood with the $\ell_1/\ell_2$ mixed norm as the penalty and we develop a scalable iterative proximal smoothing algorithm for solving the objective. On the theoretical front, we establish sufficient conditions under which the graph structure can be recovered consistently and show how to construct point-wise confidence intervals for parameters in the model. Our theoretical results are illustrated in simulation studies and the method is used to study associations between a set of stocks in the S&P 500.

## 1.1 Notation

We denote $[n]$ to be the set $\{1, \ldots, n\}$. $\mathbb{I}\{\cdot\}$ denotes the indicator function. $(x)_+ = \max(0, x)$. For a vector $a \in \mathbb{R}^n$, we let $\text{supp}(a) = \{j \ : \ a_j \neq 0\}$ be the support set (with an analogous definition for matrices $A \in \mathbb{R}^{n_1 \times n_2}$), $\|a\|_q$, $q \in [1, \infty)$, the $\ell_q$-norm defined as $\|a\|_q = (\sum_{i \in [n]} |a_i|^q)^{1/q}$ with the usual extensions for $q \in \{0, \infty\}$, that is, $\|\mathbf{a}\|_0 = |\text{supp}(\mathbf{a})|$ and $\|\mathbf{a}\|_\infty = \max_{i \in [n]} |a_i|$. For a matrix $A \in \mathbb{R}^{n_1 \times n_2}$, we use the notation $\text{vec}(A)$ to denote the vector in $\mathbb{R}^{n_1 n_2}$ formed by stacking the columns of $A$. We use the following matrix norms $\|A\|_F^2 = \sum_{i \in [n_1], j \in [n_2]} a_{ij}^2$, $\|A\|_\infty = \max_{i \in [n_1], j \in [n_2]} |a_{ij}|$ and $\|A\|_{\infty,\infty} = \max_{i \in [n_1]} \sum_{j \in [n_2]} |A_{ij}|$. The smallest and largest singular values of a matrix $A$ are denotes as $\sigma_{\min}(A)$ and $\sigma_{\max}(A)$, respectively. The characteristic function of the set of positive semidefinite matrices is denoted as $\delta_{\mathbf{S}_{++}^p}(\cdot)$ ($\delta_{\mathbf{S}_{++}^p}(A) = 0$ if $A$ is positive semidefinite and $\delta_{\mathbf{S}_{++}^p}(A) = \infty$ otherwise). For two matrices $A$ and $B$, $C = A \otimes B$ denotes the Kronecker product and $C = A \circ B$ denotes the elementwise product. For two sequences of numbers $\{a_n\}_{n=1}^\infty$ and $\{b_n\}_{n=1}^\infty$, we use $a_n = \mathcal{O}(b_n)$ to denote that $a_n \leq Cb_n$ for some finite positive constant $C$, and for all $n$ large enough. If $a_n = \mathcal{O}(b_n)$ and $b_n = \mathcal{O}(a_n)$, we use the notation $a_n \asymp b_n$. The notation $a_n = o(b_n)$ is used to denote that $a_n b_n^{-1} \xrightarrow{n \to \infty} 0$. We also use $a_n \lesssim b_n$ for $a_n = \mathcal{O}(b_n)$ and $a_n \gtrsim b_n$ for $b_n = \mathcal{O}(a_n)$.

## 2 Methodology

In this section, we propose a method for learning a sparse conditional inverse covariance matrix. Section 2.1 formalizes the problem, the estimator is given in Section 2.2, and the numerical procedure is described in Section 2.3.

## 2.1 Preliminaries

Let $\{x^i, z^i\}_{i \in [n]}$ be an independent sample from a joint probability distribution $(X, Z)$ over $\mathbb{R}^p \times [0, 1]$. We assume that the conditional distribution of $X$ given $Z = z$ is given as

$$X \mid Z = z \sim \mathcal{N}(\mu(z), \Sigma(z)). \qquad (1)$$

Let $f(z)$ be the density function of $Z$. We assume that the density is well behaved as specified later, however, we do not pose any specific distributional assumptions. Furthermore, we do not specify parametric relationships for the conditional mean or variance as functions of $Z = z$.

Our goal is to learn the conditional independence assumptions between the components of the vector $X$ given $Z = z$ and strength of associations between the components of the vector $X$ as a function of the index variable $Z$. Let $\Omega(z) = \Sigma(z)^{-1} = (\Omega_{uv}(z))_{u,v \in [p] \times [p]}$ be the inverse conditional covariance matrix. The pattern of non-zero elements of this matrix encodes the conditional independencies between the components of the vector $X$. In particular

$$X_u \perp X_v \mid X_{-uv}, Z = z \quad \text{if and only if} \quad \Omega_{uv}(z) = 0$$

where $X_{-uv} = \{X_c \mid c \in [p] \backslash \{u, v\}\}$. Denote the set of conditional dependencies given $Z = z$ as

$$S(z) = \{(u, v) \mid \Omega_{uv}(z) \neq 0\}.$$

Using $S(z)$, we define the set

$$S = \cup_{z \in [0,1]} S(z) = \{(u, v) \mid \Omega_{uv}(z) \neq 0 \text{ for some } z \in [0, 1]\}. \qquad (2)$$

Let $\bar{S}$ be the complement of $S$, which denotes pairs of components of $X$ that are conditionally independent irrespective of the value of $Z$. Under the sparsity assumption, the cardinality of set $S$ is going to be much smaller than the sample size. Our results will also depend on the maximum node degree in the graph

$$d = \max_{u \in [p]} |\{v \neq u \mid (u, v) \in S\}|.$$

In the next section, we describe our procedure for estimating $\Omega(z)$. Based on the estimator of $\Omega(z)$ we will construct an estimate of the set $S$ by reading of the positions of non-zero elements in the estimate of $\Omega(z)$.

## 2.2 Estimation Procedure

We propose to estimate the conditional sparse precision matrix using the following penalized local likelihood objective

$$\min_{\{\Omega(z^i) \succ 0\}_{i \in [n]}} \sum_{i \in [n]} \left( \text{tr}(\widehat{\Sigma}(z^i) \Omega(z^i)) - \log \det \Omega(z^i) \right) + \lambda \sum_{v \neq u} \sqrt{\sum_{i \in [n]} \Omega_{vu}(z^i)^2} \qquad (3)$$

where

$$\widehat{\Sigma}(z) = \sum_{i \in [n]} w^i(z)(x^i - \widehat{m}(z^i))(x^i - \widehat{m}(z^i))^T \quad \text{and} \quad \widehat{m}(z) = \sum_{i \in [n]} w^i(z) x^i \qquad (4)$$

are the locally weighted estimator of the covariance matrix and mean with weights given as

$$w^i(z) = \frac{K_h(z^i - z)}{\sum_{i \in [n]} K_h(z^i - z)}.$$

Here $K_h(x) = h^{-1}K(x/h)$, $K(\cdot)$ is a symmetric kernel function with bounded support, $h$ denotes the bandwidth and $\lambda$ is the penalty parameter. The penalty term

$$\lambda \sum_{v \neq u} \sqrt{\sum_{j \in [n]} (\Omega_{vu}(z^j))^2}$$

is a group sparsity regularizer (Yuan and Lin, 2006) that encourages the estimated precision matrices to have the same sparsity patterns. The parameter $\lambda > 0$ determines the strength of the penalty. Given the solution $\left\{\widehat{\Omega}(z^i)\right\}_{i \in [n]}$ of the optimization program in (3), we let

$$\widehat{S} = \left\{(u, v) \mid \widehat{\Omega}_{uv}(z^i) \neq 0 \text{ for some } i \in [n]\right\} \quad (5)$$

be the estimator of the set $S$ in (2).

Notice that in (3), we only estimate $\Omega(z)$ at $z \in \{z^i\}_{i \in [n]}$. Under the assumption that the marginal density of $Z$ is continuous and bounded away from zero on $[0, 1]$ (see Section 3) the maximal distance between any two neighboring observations is $\mathcal{O}_P(\log n/n)$ (Janson, 1987). This implies that together with the assumption that elements of $\Omega(z)$ are smooth functions on $z \in [0, 1]$, we can estimate the whole curve $\Omega_{uv}(z)$ from the point estimates $\{\widehat{\Omega}_{uv}(z^i)\}_{i \in [n]}$. The resulting approximation is of smaller order than the optimal non-parametric rate of convergence $\mathcal{O}_P(n^{-2/5})$.

**Relationship to related work.** Yin et al. (2010) proposed to estimate the parameters in (1), $\mu(z)$ and $\Sigma(z)$, by minimizing the kernel smoothed negative log-likelihood

$$\min_{\Sigma(z),\, m(z)} \sum_{i \in [n]} w^i(z) \left((x^i - m(z))\Sigma^{-1}(z)(x^i - m(z))^T - \log \det(\Sigma^{-1}(z))\right).$$

The solution to the above optimization problem is given in (4). However, these estimates work only in a low-dimensional setting when the sample size is larger than the dimension $p$. Adding the penalty in (3) allows us to estimate the parameters even when $n \ll p$.

The objective in (3) is similar to the one used in Danaher et al. (2014). Their group graphical lasso objective is

$$\min \sum_{k \in [K]} n_k \left(\text{tr}(\widehat{\Sigma}_k \Omega_k) - \log \det \Omega_k\right) + \lambda \sum_{v \neq u} \sqrt{\sum_{k \in [K]} \Omega_{k,uv}^2}$$

where $n_k$ denotes the sample size for class $k$ and $\widehat{\Sigma}_k$ is the sample covariance of observation in class $k$. Notice, however, that there are fixed number of classes, $K$, and for each class there are multiple observations that are used to estimate the class specific covariance matrix $\Sigma_k$. In contrast, we are estimating $\Sigma(z)$ at each $z \in \{z^i\}$ using the local smoothed estimator. Under the assumption that the density of $Z$ is continuous, we will have that all the observed values $\{z^i\}$ are distinct and as a result the optimization problem in (3) becomes more complicated as the sample size grows.

Finally, our proposal is related to the work of Zhou et al. (2010) and Kolar and Xing (2011) where

$$\min_{\Omega(\tau)\succ 0} \left(\text{tr}\left(\widehat{\Sigma}(\tau)\Omega(\tau)\right) - \log\det\Omega(\tau)\right) + \lambda \sum_{v\neq u} |\Omega_{vu}(\tau)| \tag{6}$$

is used to estimate the parameters at a single point $\tau \in [0,1]$ from independent observations $\{x^i \sim N(\mu(i/n), \Sigma(i/n))\}_{i\in[n]}$. Their model is different from ours in the following aspects: i) the index variable is observed on a grid $z_i = i/n$ (fixed design) and ii) the conditional independence structure can change with the index variable.

## 2.3 Proximal Iterative Smoothing

In this section, we develop a Proximal Iterative Smoothing algorithm, termed PRISMA, for solving the optimization problem in (3). We build on the work of Orabona et al. (2012) who developed a fast algorithm for finding the minimizer of the regular graphical lasso objective (Yuan and Lin, 2007, Banerjee et al., 2008, Friedman et al., 2008). The optimization problem in (3) can also be solved by the Alternating Direction Methods of Multipliers(ADMM) (Boyd et al., 2011, Danaher et al., 2014). From our numerical experience, we observed that PRISMA converges faster than ADMM for our task and is as easy as ADMM for implementation (see Figure 1 below).

The proximal iterative smoothing algorithm is suitable for solving a convex optimization that can be written as a sum of three parts: a smooth part, a simple Lipschitz part, and a non-Lipschitz part. The basic idea of PRISMA is to construct a $\beta$-Moreau envelope of the simple Lipschitz function, then perform regular accelerated proximal gradient descent on the smoothed objective. Specifically, we decompose the objective (3) into three parts: (i) the smooth part: $\text{tr}\left(\widehat{\Sigma}(z^i)\Omega(z^i)\right)$, (ii) the non-smooth Lipschitz part: $\lambda \sum_{v\neq u} \sqrt{\sum_{j\in[n]} \Omega_{vu}(z^j)^2}$, and (iii) the convex non-continuous part: $-\log\det\Omega(z^i) + \delta_{\mathbf{S}^p_{++}}(\Omega(z^i))$. Then the non-smooth Lipschitz part $\lambda \sum_{v\neq u} \sqrt{\sum_{j\in[n]} \Omega_{vu}(z^j)^2}$ is approximated by the following $\beta$-Moreau envelope

$$\varphi_\beta(\Omega(z)) = \inf_{U(z)} \left\{ \frac{\sum_i \|U(z^i) - \Omega(z^i)\|_F^2}{2\beta} + \lambda \sum_{v\neq u} \sqrt{\sum_{j\in[n]} U_{vu}(z^j)^2} \right\}.$$

The $\beta$-Moreau envelope given above is $\frac{1}{\beta}$-smooth, which means that it has a Lipschitz continuous gradient with Lipschitz constant $\frac{1}{\beta}$. We can compute the gradient with the following proximal operator

$$\nabla\varphi_\beta(\Omega(z^i)) = \frac{1}{\beta}\left(\Omega(z^i) - \text{prox}_{\ell_1\backslash\ell_2}(\Omega(z^i), \lambda\beta)\right),$$

where

$$\text{prox}_{\ell_1\backslash\ell_2}(\Omega(z^i), \lambda\beta) = \underset{\Omega}{\operatorname{argmin}} \left\{ \frac{\|\Omega - \Omega(z^i)\|_F^2}{2\lambda\beta} + \sum_{v\neq u} \sqrt{\sum_{j\in[n]} \Omega_{vu}(z^j)^2} \right\}$$

can be obtained in a closed-form (Bach et al., 2011) as

$$\text{prox}_{\ell_1 \backslash \ell_2}(\Omega(z^i), \lambda\beta)_{uv} = \Omega_{uv}(z^i) \left(1 - \frac{\lambda\beta}{\sqrt{\sum_{j\in[n]} \Omega_{vu}(z^j)^2}}\right)_+.$$

The PRISMA algorithm works by performing an accelerated proximal gradient descent on the smoothed objective

$$\sum_i \text{tr}\left(\widehat{\Sigma}(z^i)\Omega(z^i)\right) + \varphi_\beta(\Omega(z)) + \sum_i \left(-\log\det\Omega(z^i) + \delta_{\mathbf{S}^p_{++}}(\Omega(z^i))\right).$$

Let $L_f$ denote the Lispschitz constant of the gradient of $\text{tr}\left(\widehat{\Sigma}(z^i)\Omega(z^i)\right)$ and let $L_k = L_f + \frac{1}{\beta}$. Then $\text{tr}\left(\widehat{\Sigma}(z^i)\Omega(z^i)\right) + \varphi_\beta(\Omega(z))$ is $L_k$-smooth. The proximal gradient descent is performing the following update

$$\Omega(z^i) \leftarrow \text{prox}_{\left(-\log\det + \delta_{\mathbf{S}^p_{++}}\right)}\left[\Omega(z^i) - \frac{1}{L_k}\left(\nabla \text{tr}\left(\widehat{\Sigma}(z^i)\Omega(z^i)\right) + \nabla\varphi_\beta(\Omega(z))\right), \frac{1}{L_k}\right].$$

where $1/L_k$ is the step size and the proximal operator is computing a solution to the following optimization problem

$$\text{prox}_{\left(-\log\det + \delta_{\mathbf{S}^p_{++}}\right)}\left(A, \frac{1}{L_k}\right) = \underset{\Omega}{\text{argmin}}\left\{\frac{L_k\|\Omega - A\|_F^2}{2} - \log\det(\Omega) + \delta_{\mathbf{S}^p_{++}}(\Omega)\right\}. \quad (7)$$

The update can be equivalently written as

$$\Omega(z^i) \leftarrow \text{prox}_{\left(-\log\det + \delta_{\mathbf{S}^p_{++}}\right)}\left[\Omega(z^i) - \frac{1}{L_k}\left(\widehat{\Sigma}(z^i) + \frac{1}{\beta}\left(\Omega(z^i) - \text{prox}_{\ell_1 \backslash \ell_2}(\Omega(z^i), \lambda\beta)\right)\right), \frac{1}{L_k}\right].$$

Next, we describe how to compute the proximal operator in (7). Observe that (7) is equivalent to the following constrained problem

$$\min_{\Omega \succ 0} \frac{L_k\|\Omega - A\|_F^2}{2} - \log\det(\Omega).$$

From the first-order optimality condition we have that

$$L_k\Omega - \Omega^{-1} = L_k A \quad \text{and} \quad \Omega \succ 0.$$

Let $L_k A = Q\Lambda Q^T$ denote the eigen-decomposition with $\Lambda = \text{diag}(\lambda_1, \ldots, \lambda_p)$. The update will be of the form $\Omega = Q\text{diag}(\omega_1, \ldots, \omega_p)Q^T$ where $\omega_i$ can be obtained as a positive solution to the equation

$$L_k\omega_i - \frac{1}{\omega_i} = \lambda_i, \qquad i \in [p],$$

**Algorithm 1: PRISMA-CCS$_{\ell_1/\ell_2}$** — PRoximal Iterative Smoothing Algorithm for Conditional Covariance Selection with the $\ell_1/\ell_2$ penalty

**Input**: $\{(x^1, z^1), (x^2, z^2), ..., (x^n, z^n)\}$: observed data.
**Initialization**: parameters $\beta_k$, smoothness condition $L_f$, $L_k = L_f + \frac{1}{\beta_k}$, $\alpha_1 = 0$.
**for** $k = 1, 2, ...$ **do**
 $L_{k+1} \leftarrow L_f + \frac{1}{\beta_k}$
 $U_k(z) \leftarrow \Omega_k(z) - \text{prox}_{\ell_1 \setminus \ell_2}(\Omega_k(z), \beta_k \lambda)$.
 $\Theta_k(z) \leftarrow \text{prox}_{\left(-\log\det + \delta_{\mathbf{S}_{++}^p}\right)} \left(\Omega_k(z) - L_k^{-1}(\widehat{\Sigma}(z) + \beta_k^{-1} U_k(z)), 1/L_k\right)$
 $\alpha_{k+1} = \frac{1 + \sqrt{1 + 4\alpha_k^2}}{2}$
 $\Omega_{k+1}(z) = \Theta_k(z) + \frac{\alpha_k - 1}{\alpha_{k+1}}(\Theta_k(z) - \Theta_{k-1}(z))$
**end for**
**Output**: $\Omega_{k+1}(z)$

that is

$$\omega_i = \frac{\lambda_i + \sqrt{\lambda_i^2 + 4L_k}}{2L_k}. \qquad (8)$$

In summary, the proximal operator in (7), can be computed as

$$\text{prox}_{\left(-\log\det + \delta_{\mathbf{S}_{++}^p}\right)}\left(A, \frac{1}{L_k}\right) = Q\text{diag}(\omega_1, \ldots, \omega_p)Q^T$$

where $Q$ contains eigenvectors of $A$ and $\omega_i$ is given in (8).

The above described algorithm can be accelerated by combining two sequences of iterates $\Omega(z)$ and $\Theta(z)$, as in (Nesterov, 1983). The details are given in Algorithm 1.

Note that since $\text{tr}\left(\widehat{\Sigma}(z^i)\Omega(z^i)\right)$ is linear, $L_f$ can be arbitrary small. Orabona et al. (2012) suggest choosing $\beta = \mathcal{O}(1/k)$ to achieve optimal convergence $\mathcal{O}(\log k/k)$. As a rule of thumb, we found that choosing $L_f = 0.1$ and $\beta = 0.1$ gives nice convergence in our problem. Figure 1 shows a typical convergence curve of PRISMA and ADMM for our optimization problem (see Section 5 for more details on our simulation studies).

Solving (3) can be slow when the sample size is large, even with the scalable procedure described above. To speed up the estimation procedure, one may consider estimating $\Omega(z)$ at $K \ll n$ index points $\{\widetilde{z}^i\}_{i \in [K]}$, which are uniformly spread over the domain of $Z$. In this case the optimization procedure in (3) becomes

$$\min_{\{\Omega(\widetilde{z}^i) \succ 0\}_{i \in [K]}} \sum_{i \in [K]} \left(\text{tr}(\widehat{\Sigma}(\widetilde{z}^i)\Omega(\widetilde{z}^i)) - \log\det \Omega(\widetilde{z}^i)\right) + \lambda \sum_{v \neq u} \sqrt{\sum_{i \in [K]} \Omega_{vu}(\widetilde{z}^i)^2}. \qquad (9)$$

Choice of $K$ determines trade off between information aggregated across index variables and speed of computation. We report our numerical results in Section 5 using $K = 50$.

The optimization procedure can be further accelerated by adopting the covariance screening trick (Witten et al., 2011, Danaher et al., 2014) in which one identifies the connected components of the resulting graph directly from the sample covariance matrix. The optimization program in (3) is then solved for each connected component separately, which dramatically improves the running time, while still obtaining exactly the same solution as solving the whole big problem. See Section 4 in Danaher et al. (2014) for more details.

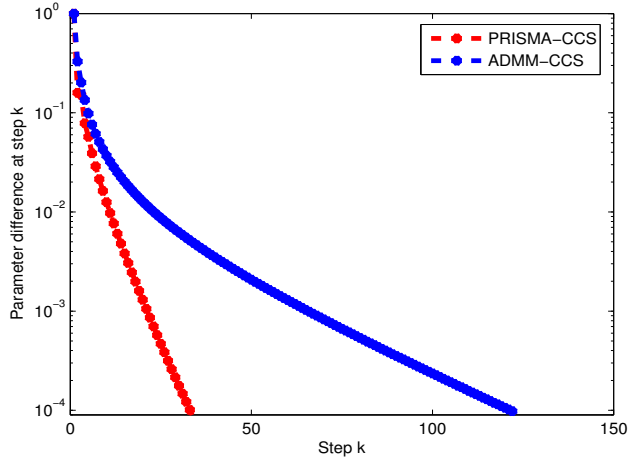

Figure 1: Typical convergence behavior of PRISMA and ADMM when solving the optimization problem in (3). The curve was obtained on a problem with $(n, p) = (500, 50)$.

## 3 Consistent Graph Identification

In this section we give a result that the set $S$ in (2) can be recovered with high-probability. We start by describing the assumption under which this result will be established. Let $A(x) = (A_{uv}(x))_{u \in [k_1], v \in [k_2]}$ with $A_{uv}(x) : \mathcal{X} \mapsto \mathbb{R}$ be a matrix of functions. Without loss of generality, let $\mathcal{X} = [0, 1]$ (otherwise we can normalize any interval $[a, b]$ to $[0, 1]$). Given a collection of index points $\mathcal{I}^n = \{x^i\}_{i \in [n]}$, let $\mathcal{H}^n(A(x)) = \mathcal{H}^n(A(x); \mathcal{I}^n) \in \mathbb{R}^{k_1 \times k_2}$ be defined as

$$\mathcal{H}^n_{uv}(A(x)) = \frac{1}{\sqrt{|\mathcal{I}^n|}} \sqrt{\sum_{x \in \mathcal{I}^n} A^2_{uv}(x)}, \quad u \in [k_1], v \in [k_2].$$

As we can see, element $(u, v)$ of $\mathcal{H}^n(A(x))$ is the quadratic mean of $A_{uv}(x^1), \ldots, A_{uv}(x^n)$. The following lemma summarizes some basic facts about $\mathcal{H}^n(\cdot)$.

**Lemma 1.** Let $A(x)$ and $B(x)$ be symmetric matrices for all $x \in \mathcal{X}$. Given $n$ index points in $\mathcal{X}$, $\{x^i\}_{i \in [n]}$, we have

(a) $\|\mathcal{H}^n(A(x))\|_{\infty,\infty} \leq \max_{i \in [n]} \|A(x^i)\|_{\infty,\infty}$

(b) $\|\mathcal{H}^n(A(x))\|_{\infty} \leq \max_{i \in [n]} \|A(x^i)\|_{\infty}$

(c) $\|\mathcal{H}^n(A(x)B(x))\|_{\infty,\infty} \leq \|\mathcal{H}^n(A(x))\|_{\infty,\infty} \|\mathcal{H}^n(B(x))\|_{\infty,\infty}$

(d) $\|\mathcal{H}^n(A(x)B(x))\|_{\infty} \leq \|\mathcal{H}^n(A(x))\|_{\infty} \|\mathcal{H}^n(B(x))\|_{\infty,\infty}$

(e) $\|\mathcal{H}^n(A(x) + B(x))\|_\infty \leq \|\mathcal{H}^n(A(x)) + \mathcal{H}^n(B(x))\|_\infty.$

We will use the following assumptions to develop our theoretical results.

**A1 (Distribution).** We assume the data $\{(x^i, z^i)\}_{i \in [n]}$ are drawn i.i.d. from the following model

$$\begin{aligned} z^i &\sim f(z) \\ x^i \mid z^i &\sim \mathcal{N}(0, \Sigma^*(z^i)). \end{aligned} \tag{10}$$

Marginal density $f(z)$ of $Z$ is lower bounded on the support $[0, 1]$, that is, there is a constant $C_f$ such that

$$\inf_{z \in [0,1]} f(z) \geq C_f > 0.$$

Furthermore, the density function is $C_d$-smooth, that is, the first derivative is $C_d$-Lipschitz

$$|f'(z) - f'(z')| \leq C_d |z - z'|, \qquad \forall z, z' \in [0, 1].$$

**A2 (Kernel).** The kernel function $K(\cdot) : \mathbb{R} \to \mathbb{R}$ is a symmetric, probability density function supported on $[-1, 1]$. There exists a constant $C_K < \infty$, such that

$$\sup_x |K(x)| \leq C_K \quad \text{and} \quad \sup_x K(x)^2 \leq C_K.$$

Furthermore, the kernel function is $C_L$-Lipschitz on the support $[-1, 1]$, that is

$$|K(x) - K(x')| \leq C_L |x - x'|, \qquad \forall x, x' \in [-1, 1].$$

**A3 (Covariance).** There exist constants $\Lambda_\infty, C_\infty < \infty$, such that

$$\frac{1}{\Lambda_\infty} \leq \inf_z \sigma_{\min}(\Sigma^*(z)) \leq \sup_z \sigma_{\max}(\Sigma^*(z)) \leq \Lambda_\infty$$

and

$$\sup_z \|\Sigma^*(z)\|_{\infty,\infty} \leq C_\infty.$$

**A4 (Smoothness).** There exists a constat $C_\Sigma$, such that

$$\max_{u,v} \sup_z \left|\frac{d}{dz}\Sigma^*_{uv}(z)\right| \leq C_\Sigma \quad \text{and} \quad \max_{u,v} \sup_z \left|\frac{d^2}{dz^2}\Sigma^*_{uv}(z)\right| \leq C_\Sigma$$

$$\max_u \sup_z \left|\frac{d}{dz}m^*_u(z)\right| \leq C_\Sigma \quad \text{and} \quad \max_u \sup_z \left|\frac{d^2}{dz^2}m^*_u(z)\right| \leq C_\Sigma.$$

**A5 (Irrepresentable Condition).** Let

$$\mathcal{I}(z) = \Omega^*(z)^{-1} \otimes \Omega^*(z)^{-1}.$$

The sub-matrix $\mathcal{I}_{SS}(z)$ is invertible for all $z \in [0,1]$. There exist constants $\alpha \in (0,1]$ and $C_\mathcal{I}$ such that

$$\sup_z \|\|\mathcal{I}_{S^cS}(z)(\mathcal{I}_{SS}(z))^{-1}\|\|_{\infty,\infty} \leq 1 - \alpha \quad \text{and} \quad \sup_z \|\|\mathcal{I}_{SS}(z)\|\|_{\infty,\infty} \leq C_\mathcal{I}.$$

Assumption A1 on the density function $f(z)$ is rather standard (see, for example, Li and Racine, 2006). In (10), we assume that the conditional mean $\mu(z) = E[X \mid Z = z]$ is equal to zero. This is assumption is not necessary, however, it makes for simpler exposition and derivation of results. Alternatively, we could impose a smoothness condition on the condtional mean, for example, that each component of $m(\cdot)$ has two continuous derivatives. This alternative assumption would not change results presented in this and the following section as the rate of convergence will be dominated by how fast we can estimate elements of $\Omega^*(z)$. The assumption that $X \mid Z$ follows a multivariate Normal distribution is quite strong. This could be changed to a sub-Gaussian distribution and our results would not change except for constants. However, establishing our results under moment conditions would require different proof techniques. Assumption A2 is a standard technical condition that is used in literature on kernel smoothing (see, for example, Li and Racine, 2006). A number of commonly used kernel function satisfy A2, such as the Box car kernel $K(x) = \frac{1}{2}\mathbb{I}(|x| \leq 1)$, the Epanechnikov kernel $K(x) = \frac{3}{4}(1-x^2)\mathbb{I}(|x| \leq 1)$ and the Tricube kernel $K(x) = \frac{70}{81}(1-|x|^3)\mathbb{I}(|x| \leq 1)$. Assumption A3 ensures that the model is identifiable in the population setting. Our assumption could be relaxed by allowing $\Lambda_\infty$ and $C_\infty$ to depend on the sample size $n$ as in Ravikumar et al. (2011), however, we choose not do keep track of this dependence to simplify the exposition. Assumption A4 ensures sufficient smoothness of the mean and covariance functions. These conditions conditions are necessary (Yin et al., 2010). Assumption A5 is a version of the irrepresentable condition (Zhao and Yu, 2006, Wainwright, 2009, Meinshausen and Bühlmann, 2006) commonly used in the high-dimensional literature to ensure exact model selection. We use an assumption on the Fisher information matrix previously imposed in Ravikumar et al. (2011), however, we assume that it holds uniformly for all values of $z \in [0,1]$.

**Remark 2.** *It follows directly from Lemma 1 that given the index points $\{z^i\}_{i \in [n]}$, we have*

$$\|\mathcal{H}^n(\Sigma^*(z))\|_{\infty,\infty} \leq \sup_z \|\Sigma^*(z)\|_{\infty,\infty} \leq C_\infty,$$

$$\|\mathcal{H}^n(\mathcal{I}_{S^cS}(z)(\mathcal{I}_{SS}(z))^{-1})\|_{\infty,\infty} \leq \sup_z \|\mathcal{I}_{S^cS}(z)(\mathcal{I}_{SS}(z))^{-1}\|_{\infty,\infty} \leq 1 - \alpha, \text{ and}$$

$$\|\mathcal{H}^n(\mathcal{I}_{SS}(z))\|_{\infty,\infty} \leq \sup_z \|\mathcal{I}_{SS}(z)\|_{\infty,\infty} \leq C_\mathcal{I}.$$

Our first result provides conditions needed for correct recovery of the sparsity pattern of the conditional inverse covariance matrix.

**Theorem 3.** *Suppose that assumptions A1-A5 hold. Let $\{x^i, z^i\}_{i \in [n]}$ be independent observations from the model in (10). Suppose there are two constants $C_1$ and $C_2$ such that*

$$\overline{\Omega}_{\min} = \min_{(u,v) \in S} \sqrt{n^{-1} \sum_{i \in [n]} (\Omega^*_{uv}(z^i))^2} \geq C_1 n^{-2/5} \sqrt{\log p}$$

and the sample size satisfies
$$n \geq C_2 d^{5/2} (\log p)^{5/4}.$$
If the bandwidth is chosen as $h \asymp n^{-1/5}$ and the penalty parameter as $\lambda \asymp n^{1/10} \sqrt{\log p}$, then
$$\mathrm{pr}(\widehat{S} = S) = 1 - \mathcal{O}(p^{-1}) \xrightarrow{n \to \infty} 1,$$
where $\widehat{S}$ is the estimated support given in (5) and $S$ is given in (2).

The proof of Theorem 3 is given in Appendix A.1 and, at a high-level, is based on the primal-dual witness proof described in Ravikumar et al. (2011). The main technical difficulty lies in establishing a bound on $\sup_{z \in [0,1]} \max_{u,v} \left| \widehat{\Sigma}_{uv}(z) - \Sigma^*_{uv}(z) \right|$, which is provided in Lemma 8. This result is new and may be of independent interest.

Our results should be compared to those in Kolar and Xing (2011), where $\widehat{S}(z)$ was estimated using the optimization problem in (6). Kolar and Xing (2011) required the sample size $n \geq Cd^3 (\log p)^{3/2}$, while our procedure requires the sample size to satisfy $n \geq Cd^{5/2} (\log p)^{5/4}$. This difference primarily arises from more careful treatment of the smoothness of the component functions of $\Omega^*(z)$. Next, we compare requirements on the minimum signal strength (commonly referred to as the $\beta_{\min}$ condition). Kolar and Xing (2011) require that
$$\min_{(u,v) \in S(z)} |\Omega^*_{uv}(z)| \geq C n^{-1/3} \sqrt{\log p} \qquad z \in [0,1],$$
while our procedure relaxes this condition to
$$\min_{(u,v) \in S} \sqrt{n^{-1} \sum_{i \in [n]} (\Omega^*_{uv}(z^i))^2} \geq C n^{-2/5} \sqrt{\log p}.$$
Furthermore, we only require the following functionals of the covariance and Fisher information matrix to be upper bounded
$$\|\mathcal{H}^n(\Sigma^*(z))\|_{\infty,\infty}, \quad \|\mathcal{H}^n(\mathcal{I}_{S^cS}(z)(\mathcal{I}_{SS}(z))^{-1})\|_{\infty,\infty}, \quad \text{and} \quad \|\mathcal{H}^n(\mathcal{I}_{SS}(z))\|_{\infty,\infty},$$
while Kolar and Xing (2011) required the conditions to hold uniformly in $z$, which is a more stringent condition as shown in Remark 2.

**Remark 4.** *As discussed in Section 2.3, the set $\widehat{S}$ can be estimated using the optimization program in (9) instead of the one in (3). The statement of Theorem 3 remains to hold if*
$$\min_{(u,v) \in S} \sqrt{K^{-1} \sum_{i \in [K]} (\Omega^*_{uv}(z^i))^2} \geq C_1 n^{-2/5} \sqrt{\log p}.$$

The following proposition gives a consistency result on the estimation of $\Omega^*(z)$ in Frobenius norm.

**Proposition 5.** *Under the assumption of Theorem 3, there exists a constant $C$, such that*
$$n^{-1} \sum_{i \in [n]} \|\widehat{\Omega}(z^i) - \Omega^*(z^i)\|_F^2 \leq C n^{-4/5} (|S| + p) \log p$$
*with probability $1 - \mathcal{O}(p^{-1})$.*

The proof of Proposition 5 is given in Appendix A.2. Zhou et al. (2010) obtained $\mathcal{O}_P \left( n^{-2/3} (|S| + p) \log n \right)$ as the rate of convergence in the Frobenius norm. The improvement in the rate, from $n^{-2/3}$ to $n^{-4/5}$ is due to a more careful treatment of the smoothness of the precision matrix functions.

# 4 Construction of Confidence Intervals for Edges

In this section we describe how to construct point-wise confidence intervals for individual elements of $\Omega^*(z)$. Our approach is based on the idea presented in (Jankova and van de Geer, 2014) where confidence intervals are constructed for individual elements of the inverse covariance matrix based on the de-sparsified estimator and its asymptotic Normality.

Let $\{\widehat{\Omega}(z^i)\}_{i \in [n]}$ be the minimizer of (3). Based on this minimizer, we construct the following de-sparsified estimator

$$\widehat{T}(z) = \text{vec}\left(\widehat{\Omega}(z)\right) - \left(\widehat{\Omega}(z) \otimes \widehat{\Omega}(z)\right) \text{vec}\left(\widehat{\Sigma}(z) - \widehat{\Omega}(z)^{-1}\right), \quad z \in [0,1]. \tag{11}$$

The second term in (11) is to remove the bias in $\widehat{\Omega}(z)$ introduced by the sparsity inducing penalty. We have the following theorem about $\widehat{T}(z)$.

**Theorem 6.** *Suppose that the conditions of Theorem 3 are satisfied. For a fixed $z \in [0,1]$, we have*

$$\frac{n^{2/5}\sqrt{\widehat{f}(z)}\left(\widehat{T}_{uv}(z) - \Omega^*_{uv}(z)\right)}{\sqrt{\left(\widehat{\Omega}^2_{uv}(z) + \widehat{\Omega}_{uu}(z)\widehat{\Omega}_{vv}(z)\right) \int_{-\infty}^{\infty} K^2(t) dt}} = N_{uv}(z) + bias + o_P(1)$$

*where $N_{uv}(z)$ converges in distribution to $\mathcal{N}(0,1)$, and $bias = \mathcal{O}(1)$.*

The proof of theorem is given in Appendix A.3. An explicit form for the bias term can be found in the proof. The bias term can be removed by under-smoothing, that is, using a smaller bandwidth. We have the following corollary.

**Corollary 7.** *Suppose the bandwidth satisfies $h \asymp n^{-1/4}$, the penalty $\lambda \asymp n^{1/8}\sqrt{\log p}$, and sample size satisfies $n \gtrsim d^{8/3}(\log p)^{4/3}$. If*

$$\min_{u,v \in S} \sqrt{n^{-1} \sum_{i \in [n]} \Omega^*_{uv}(z^i)} \geq C n^{-3/8} \sqrt{\log p}$$

*for some constant $C$, then*

$$\frac{n^{3/8}\sqrt{\widehat{f}(z)}\left(\widehat{T}_{uv}(z) - \Omega^*_{uv}(z)\right)}{\sqrt{\left(\widehat{\Omega}^2_{uv}(z) + \widehat{\Omega}_{uu}(z)\widehat{\Omega}_{vv}(z)\right) \int_{-\infty}^{\infty} K^2(t) dt}} = N_{uv}(z) + o_P(1)$$

*where $N_{uv}(z)$ converges in distribution to $\mathcal{N}(0,1)$.*

Based on the theorem and its corollary, we can construct $100(1-\alpha)\%$ asymptotic confidence intervals $B_{uv}(z)$ for $\Omega^*_{uv}(z)$ as

$$B_{uv}(z) = \left[\widehat{T}_{uv}(z) - \delta(z,\alpha,n), \widehat{T}_{uv}(z) + \delta(z,\alpha,n)\right]$$

where

$$\delta(z,\alpha,n) = \Phi^{-1}(1-\alpha/2) n^{-3/8} \sqrt{\frac{\widehat{\Omega}^2_{uv}(z) + \widehat{\Omega}_{uu}(z)\widehat{\Omega}_{vv}(z)}{\widehat{f}(z)} \int_{-\infty}^{\infty} K^2(t) dt}.$$

We will illustrate finite sample properties of the confidence interval in Section 5.

# 5  Simulation Studies

In this section we present extensive simulation results. First, we report results on recovery of graph structure and precision matrix. Next, we report coverage properties and average length of the confidence intervals. These simulation studies serve to illustrate finite sample performance of our method. We compare against two other methods: 1) a method that ignores the index variable $Z$ and uses the glasso (Yuan and Lin, 2007) to estimate the underlying conditional independence structure, and 2) a method from Zhou et al. (2010) which does not exploit the common structure across the indexing variable. We denote the first procedure as the *glasso*, the second one as the *locally smoothed glasso*. For the locally smoothed glasso and our procedure, we use the Epanechnikov kernel defined as

$$K_h(x) = \frac{3}{4}\left(1 - \left(\frac{x}{h}\right)^2\right)\mathbb{I}\{|x| \leq h\}.$$

For our procedure, we solve the optimization problem in (9) at index points $\{0, 0.02, \ldots, 1\}$.

We first explain the experimental setup. Data are generated according to the model in (10), for which we need to specify the conditional independence structure given a graph and the edge values as functions of the index variable. We generate four kinds of graphs, described below.

- Chain graph is a random graph formed by permuting the nodes and connecting them in succession.

- Nearest Neighbor graph is a random graph generated by the procedure described in Li and Gui (2006). For each node, a point is drawn uniformly at random on a unit square. Then pairwise distances between nodes are computed and each node is connected to $d = 2$ closest neighbors.

- Erdös-Rényi graph is a random graph where each pair of nodes is connected independently of other pairs. We generated a random graph with $2p$ edges and maximum degree of the nodes is set to 5 (see, for example, Meinshausen and Bühlmann, 2006).

- Scale-free graph is a random graph created according to the procedure described in Barabási and Albert (1999). The network begins with an initial 5 nodes clique. New nodes are added to the graph one at a time and each new node is connected to one of the existing nodes with a probability proportional to the node's current degree. Formally, the probability $p_i$ that the new node is connected to an existing node $i$ is $p_i = \frac{d_i}{\sum d_i}$ where $d_i$ is the degree of node $i$.

Given the graph, we generate a precision matrix with the sparsity pattern adhering to the graph structure, that is, $\Omega^*_{uv}(z) \equiv 0$ if $(u, v) \notin S$. Non-zero components of $\Omega^*(z)$ were generated in one of the following ways:

- Random walk function. First, we generate $\widetilde{\Omega}_{uv}(0)$ uniformly at random from $[-0.3, -0.2] \cup [0.2, 0.3]$. Next, we set

$$\widetilde{\Omega}_{uv}(t/T) = \widetilde{\Omega}((t-1)/T) + (1 - 2*\text{Bern}(1/2))*0.002$$

  where $T = 10^4$. Finally, $\Omega^*_{uv}(z)$ is obtained by smoothing $\widetilde{\Omega}_{uv}(z)$ with a cubic spline.

- Linear function. We generate $\Omega^*_{uv}(z) = 2z - c_{uv}$ where $c_{uv} \sim \text{Uniform}([0,1])$ is a random offset.

- Sin function. We generate $\Omega^*_{uv}(z) = \sin(2\pi z + c_{uv})$ where $c_{uv} \sim \text{Uniform}([0,1])$ is a random offset.

To ensure the positive-definiteness of $\Omega^*(z)$ we add $\delta(z)I$ to $\Omega^*(z)$, so that the minimum eigenvalue of $\Omega^*(z)$ is at least 0.5.

In the simulation studies, we change the sample size $n$ and the number of variables $p$ as:i) $n = 500, p = 20$, ii) $n = 500, p = 50$, and iii) $n = 500, p = 100$. Once the model is fixed, we generate 10 independent data sets according to the model and report the average performance over these 10 runs.

Figure 2 shows the Precision-Recall curves for different estimation methods when the non-zero components of $\Omega^*(z)$ are generated as random walk functions. For point smoothing glasso, we report the results for graph estimated using $z = 0.5$. The precision and recall of the estimated graph are given as

$$\text{Precision}(\widehat{S}, S) = 1 - \frac{|\widehat{S} \setminus S|}{|\widehat{S}|} \quad \text{and} \quad \text{Recall}(\widehat{S}, S) = 1 - \frac{|S \setminus \widehat{S}|}{|S|},$$

where $\widehat{S}$ is the set of estimated edges. Each row in the figure corresponds to a different graph type, while columns correspond to different sample and problem sizes. Figures 3 and 4 similarly show Precision-Recall curves for linear and sin component functions. Results are further summarized in Figure 5 which shows $F_1$ score for the graph estimation task, given as the harmonic mean of precision and recall

$$F_1 = 2 \cdot \frac{\text{Precision} \cdot \text{Recall}}{\text{Precision} + \text{Recall}}.$$

Finally, Figure 6 shows the performance of different procedures in terms of squared Frobenius error $n^{-1} \sum_{i \in [n]} \|\widehat{\Omega}(z^i) - \Omega^*(z^i)\|_F^2$.

From these results, several observations could be drawn:

- For precision matrix functions: random walk, linear and sin, our procedure performs significantly better than naive graphical lasso in terms of precision-recall curve, F1, and Frobenius norm error. For example, under the sin precision function case, for the chain and NN graph, our procedure can recover the graph perfectly while the glasso totally fails. When comparing the glasso and point smoothing glasso, we found that glasso performs better for random walk function, while point smoothing glasso performs better for sin function.

- When comparing results for different types of graphs, we found that the chain and NN graphs are relatively easy to recover, and Scale-free networks are the hardest to recover due to presence of high-degree nodes. Recall that for chain and NN graph the maximum degree is small, for Erdös-Rényi graph the maximum degree is larger ($d \approx 5$), and for scale-free graph the maximum degree is the largest.

As the theory suggests, $\Omega(d^{5/2}(\log p)^{5/4})$ sample is needed for consistent graph recovery. To empirically study this behavior, we vary the sample size $n$ as $n = C \cdot d^{5/2}(\log p)^{5/4}$ by varying $C$, and plot the hamming distance between the true and estimated graph against $C$, and the theory suggest that the hamming distance curves are going to reach zero for different $p$ at a same point. Figure 7 shows the plot, and we can see that generally this is true for all the graphs and precision matrix function we studied here.

In the next set of experiments our purpose is to verify the asymptotic normality of our de-biased estimator. We follow the simulations done in Jankova and van de Geer (2014). Using Theorem 6 we construct $100(1-\alpha)\%$ confidence intervals as

$$B_{uv}(z^{(i)}) := [\widehat{T}_{uv}(z^{(i)}) - \delta(\alpha, n), \widehat{T}_{uv}(z^{(i)}) + \delta(z^{(i)}, \alpha, n)],$$

where

$$n^{-3/8}\delta(z^{(i)}, \alpha, n) := \Phi^{-1}(1-\alpha/2)\sqrt{\frac{\widehat{\Omega}_{uv}^2(z^{(i)}) + \widehat{\Omega}_{uu}(z^{(i)})\widehat{\Omega}_{vv}(z^{(i)})}{\widehat{f}(z)} \int_{-\infty}^{\infty} K^2(t)\mathrm{d}t}$$

For the simulation studies, we follow the same graph and precision construction described earlier, set $\alpha = 0.025$, and set the bandwidth $h = n^{-1/4}$ and the regularization parameter $\lambda = n^{-3/8}\sqrt{\log(p)}$ as the theory suggests. We fix the model and generate different data sets 100 times, test the empirical frequency that $\Omega^*(z^i)$ is covered by the constructed confidence intervals:

$$\widehat{\alpha}_{uv} = \frac{1}{K}\sum_{i=1}^{K}\frac{1}{100}\sum_{t=1}^{100}\mathbb{I}_{\{\Omega_{uv}^*(z^{(i)}) \in B_{uv}^t(z^{(i)})\}}$$

where $B_{uv}^t(z^{(i)})$ is the confidence intervals built based on $t$-th data set. And the average coverage on support $S$ is defined as

$$\mathrm{Avgcov}_S = \frac{1}{|S|}\sum_{(u,v)\in S}\widehat{\alpha}_{uv}.$$

Likewise, we can define the $\mathrm{Avgcov}_{S^c}$, as well the average length of the confidence intervals $\mathrm{Avglength}_S$, $\mathrm{Avglength}_{S^c}$.

Table 1, 2, 3 shows coverage properties for random walk, linear, and sin component functions, respectively. We make the following observations:

- When looking at the Avgcov, confidence intervals cover $\Omega^*(z^i)$ reasonably well, no matter what kinds of graph, for what kind of precision matrix function, which verifies the theories of asymptotic normality.

- When looking at the Avglength, and comparing different types of graph structures, we can see that the Avglength for chain and NN graphs are much shorter than the that of Erdös-Rényi and scale-free graph, because generally the later are harder to estimate.

Figure 8 illustrates confidence bands constructed by our procedure on a $n = 500, p = 50$ chain graph problem.

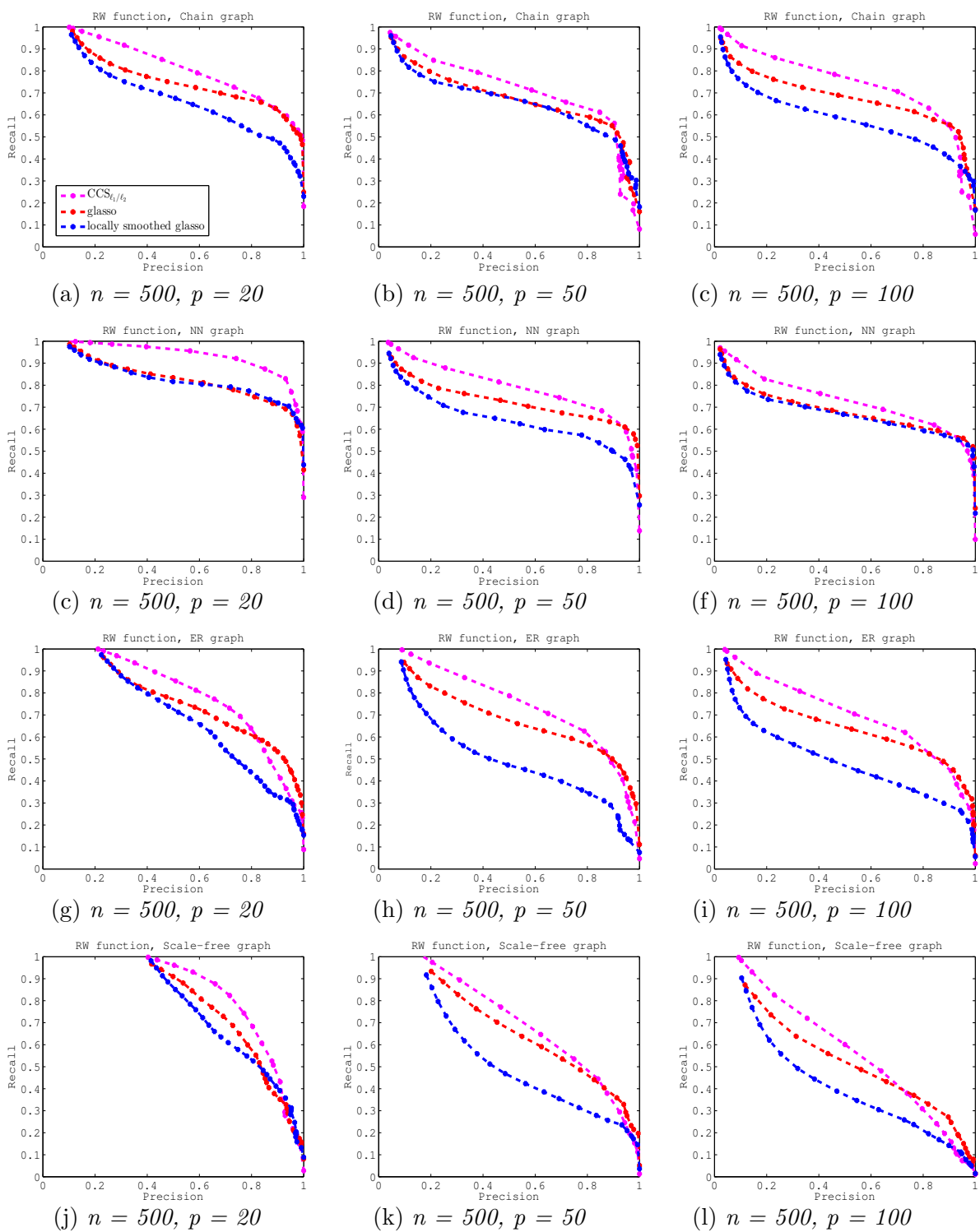

Figure 2: Precision-Recall curves for random walk function, from top to bottom: chain graph, NN graph, Erdös-Rényi graph, Scale-free graph.

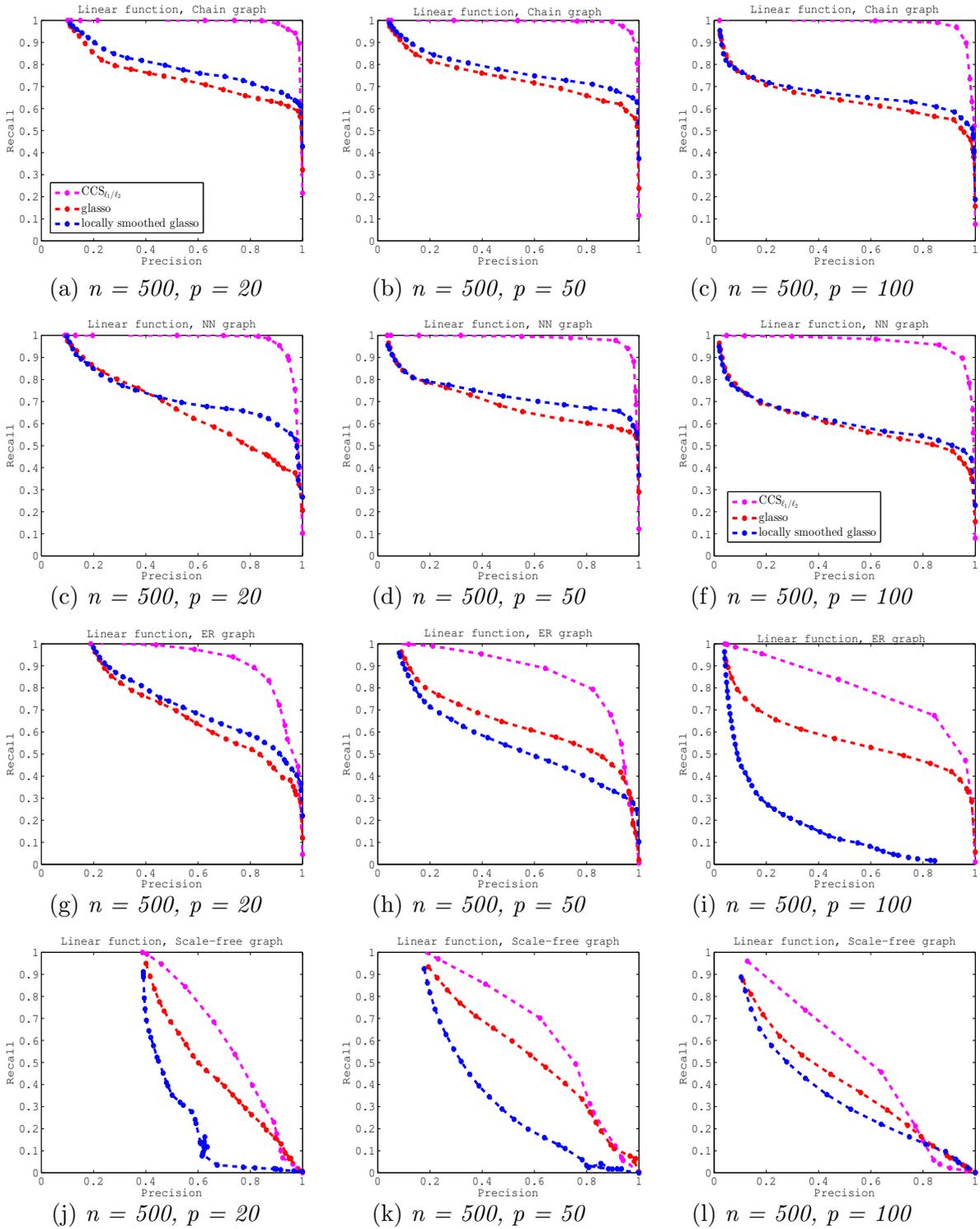

Figure 3: Precision-Recall curves for linear function, from top to bottom: chain graph, NN graph, Erdös-Rényi graph, Scale-free graph.

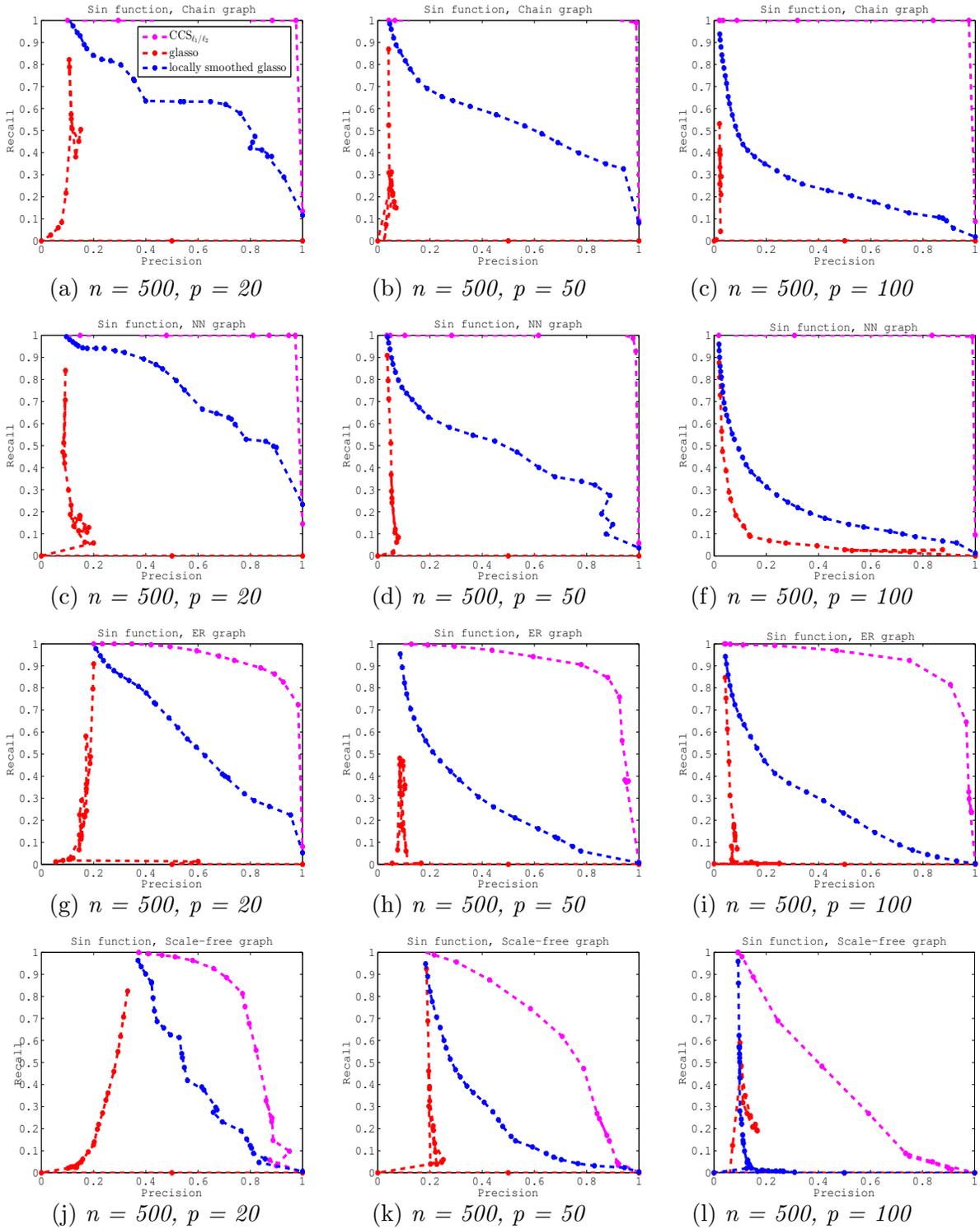

Figure 4: Precision-Recall curves for random sin function, from top to bottom: chain graph, NN graph, Erdös-Rényi graph, Scale-free graph.

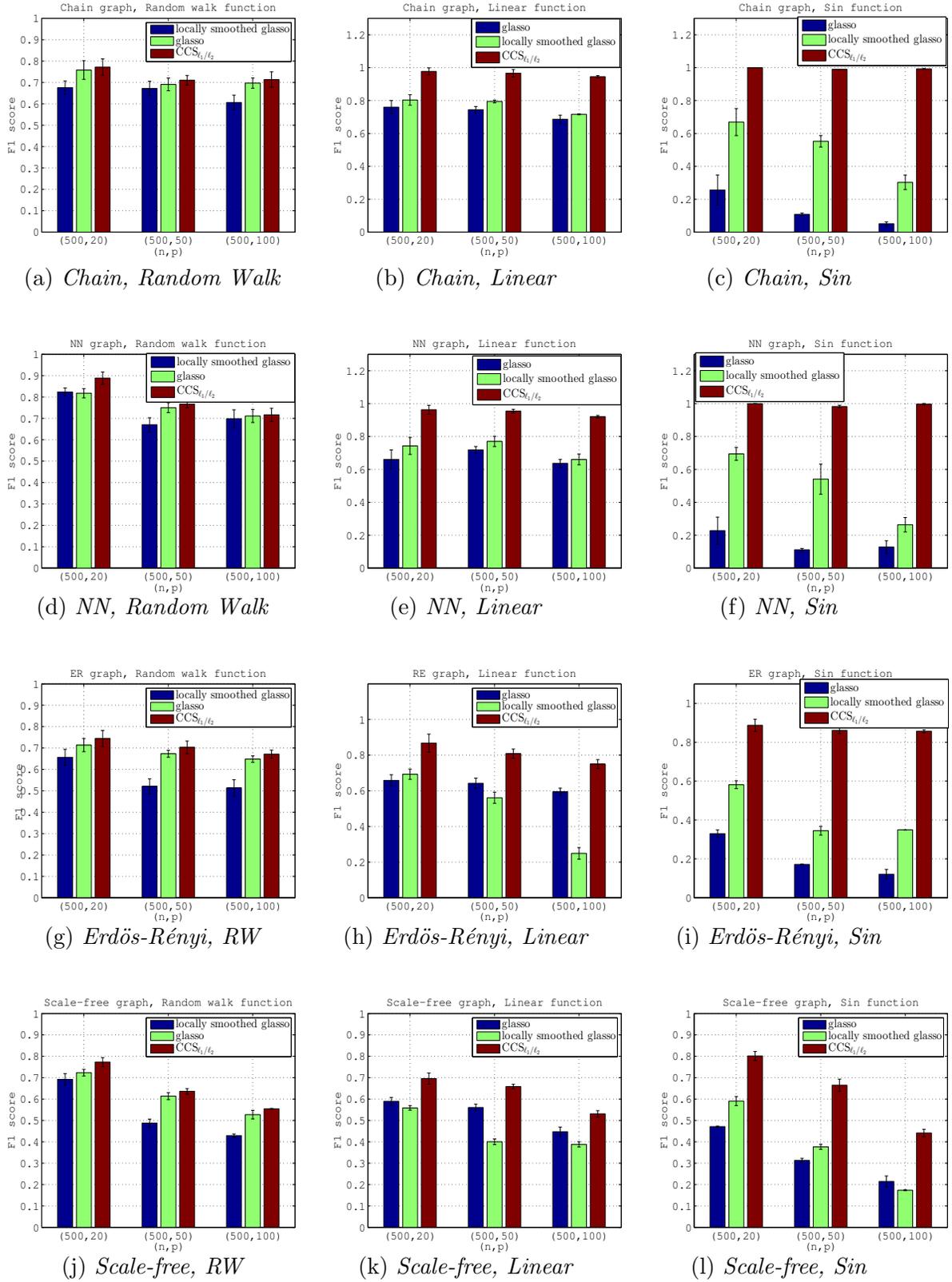

Figure 5: Average F1-scores of various methods.

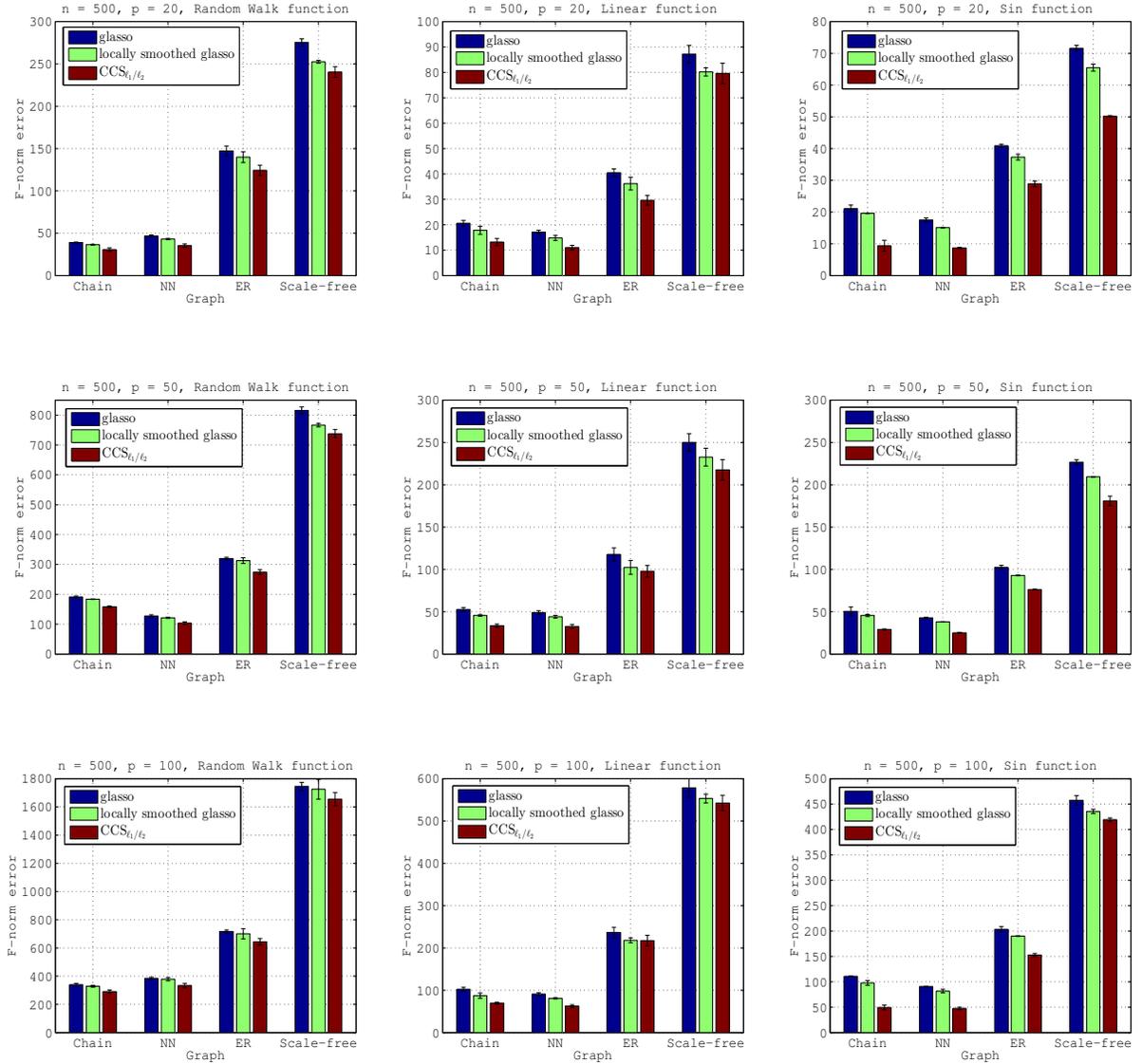

Figure 6: Average F-norm errors of various methods.

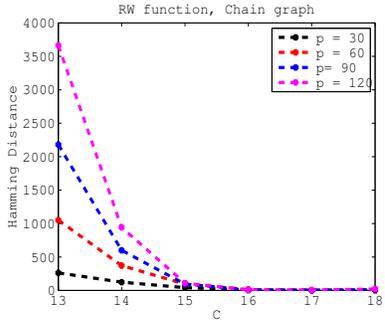
(a) *Chain, Random Walk*

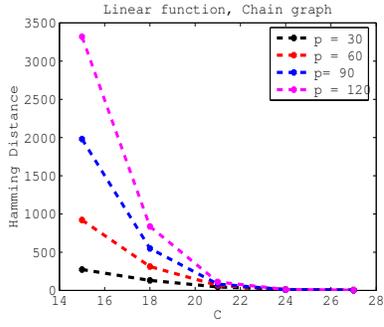
(b) *Chain, Linear*

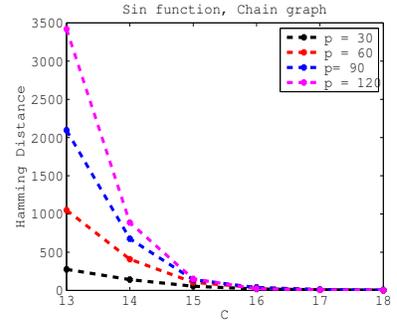
(c) *Chain, Sin*

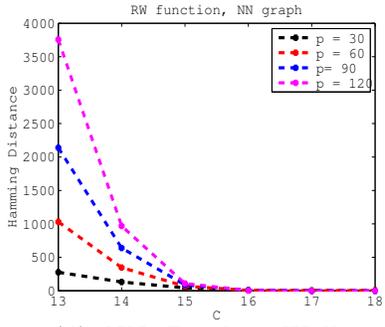
(d) *NN, Random Walk*

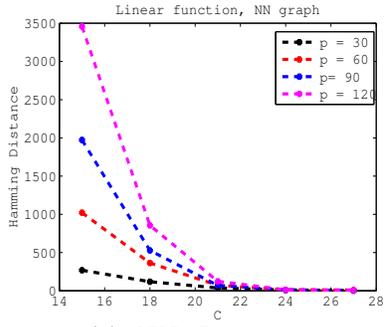
(e) *NN, Linear*

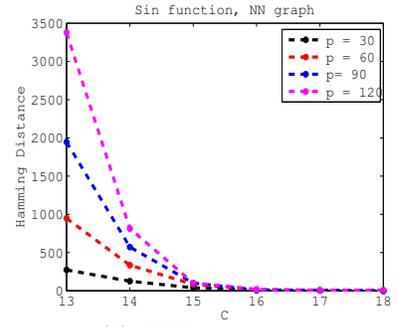
(f) *NN, Sin*

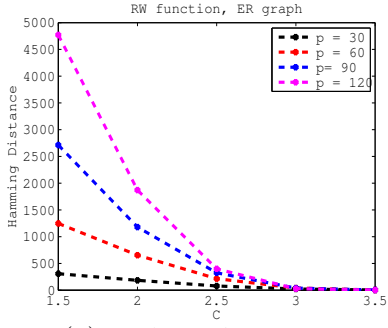
(g) *Erdös-Rényi, RW*

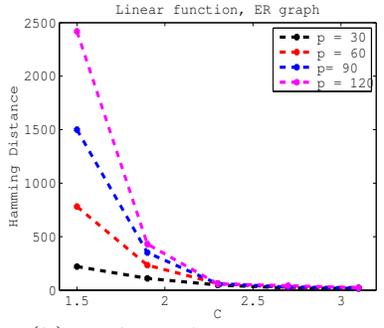
(h) *Erdös-Rényi, Linear*

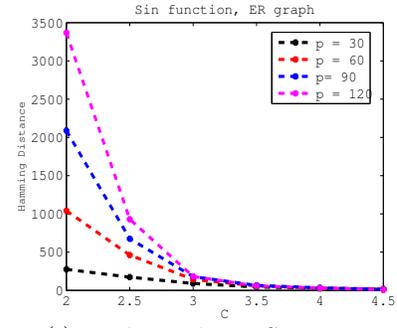
(i) *Erdös-Rényi, Sin*

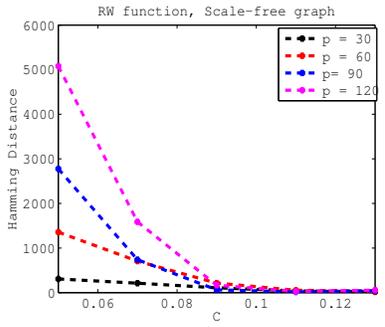
(j) *Scale-free, RW*

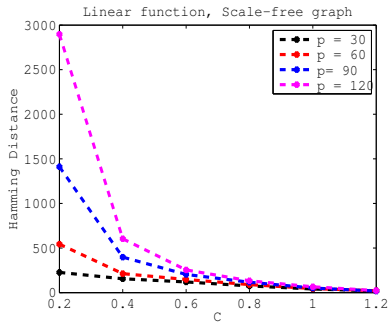
(k) *Scale-free, Linear*

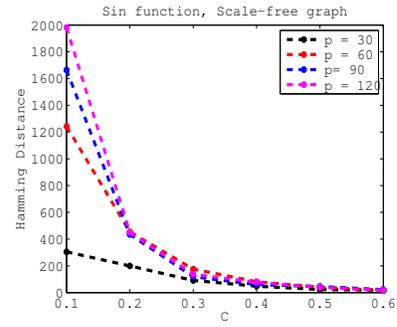
(l) *Scale-free, Sin*

Figure 7: Average hamming distance plotted against the re-scaled sample size.

| Graph(n,p) | Avgcov | | | | Avglength | | | |
|---|---|---|---|---|---|---|---|---|
| | S | $S^c$ | $(u,v) \in S$ | $(u,v) \in S^c$ | S | $S^c$ | $(u,v) \in S$ | $(u,v) \in S^c$ |
| Chain(500,20) | 0.966 | 0.973 | 0.950 | 0.965 | 0.422 | 0.413 | 0.428 | 0.416 |
| Chain(500,50) | 0.982 | 0.973 | 0.927 | 0.972 | 0.490 | 0.484 | 0.497 | 0.484 |
| Chain(500,100) | 0.973 | 0.959 | 0.881 | 0.936 | 0.547 | 0.544 | 0.540 | 0.525 |
| NN(500,20) | 0.961 | 0.972 | 0.930 | 0.980 | 0.410 | 0.403 | 0.408 | 0.408 |
| NN(500,50) | 0.976 | 0.972 | 0.935 | 0.945 | 0.495 | 0.491 | 0.496 | 0.474 |
| NN(500,100) | 0.974 | 0.960 | 0.918 | 0.981 | 0.531 | 0.529 | 0.544 | 0.521 |
| Erdös-Rényi(500,20) | 0.974 | 0.974 | 0.993 | 0.943 | 0.518 | 0.515 | 0.518 | 0.506 |
| Erdös-Rényi(500,50) | 0.981 | 0.977 | 0.945 | 0.963 | 0.614 | 0.616 | 0.617 | 0.614 |
| Erdös-Rényi(500,100) | 0.982 | 0.972 | 0.963 | 0.981 | 0.646 | 0.647 | 0.669 | 0.665 |
| Scale-free(500,50) | 0.973 | 0.971 | 0.997 | 0.986 | 0.622 | 0.619 | 0.610 | 0.619 |
| Scale-free(500,50) | 0.982 | 0.980 | 0.961 | 0.986 | 0.754 | 0.765 | 0.707 | 0.682 |
| Scale-free(500,100) | 0.982 | 0.980 | 0.927 | 0.990 | 0.808 | 0.818 | 0.777 | 0.798 |

Table 1: Confidence interval, for random walk function precision

| Graph(n,p) | Avgcov | | | | Avglength | | | |
|---|---|---|---|---|---|---|---|---|
| | S | $S^c$ | $(u,v) \in S$ | $(u,v) \in S^c$ | S | $S^c$ | $(u,v) \in S$ | $(u,v) \in S^c$ |
| Chain(500,20) | 0.954 | 0.982 | 0.926 | 0.979 | 0.775 | 0.760 | 0.782 | 0.772 |
| Chain(500,50) | 0.962 | 0.988 | 0.970 | 1 | 0.845 | 0.832 | 0.852 | 0.823 |
| Chain(500,100) | 0.960 | 0.990 | 0.919 | 1 | 0.853 | 0.840 | 0.870 | 0.815 |
| NN(500,20) | 0.956 | 0.982 | 0.934 | 0.995 | 0.721 | 0.706 | 0.740 | 0.706 |
| NN(500,50) | 0.955 | 0.987 | 0.936 | 0.991 | 0.815 | 0.802 | 0.828 | 0.775 |
| NN(500,100) | 0.947 | 0.989 | 0.946 | 0.970 | 0.819 | 0.812 | 0.838 | 0.828 |
| Erdös-Rényi(500,20) | 0.954 | 0.983 | 0.920 | 0.991 | 1.083 | 1.087 | 1.086 | 1.103 |
| Erdös-Rényi(500,50) | 0.948 | 0.990 | 0.914 | 0.995 | 1.104 | 1.112 | 1.112 | 1.107 |
| Erdös-Rényi(500,100) | 0.955 | 0.988 | 0.921 | 1 | 1.146 | 1.158 | 1.131 | 1.164 |
| Scale-free(500,20) | 0.890 | 0.978 | 0.941 | 0.960 | 1.455 | 1.473 | 1.371 | 1.431 |
| Scale-free(500,50) | 0.932 | 0.990 | 0.941 | 0.988 | 1.733 | 1.768 | 1.674 | 1.655 |
| Scale-free(500,100) | 0.947 | 0.993 | 0.946 | 0.980 | 2.011 | 2.069 | 1.810 | 1.808 |

Table 2: Confidence interval, for linear function precision

| Graph(n,p) | Avgcov | | | | Avglength | | | |
|---|---|---|---|---|---|---|---|---|
| | S | $S^c$ | $(u,v) \in S$ | $(u,v) \in S^c$ | S | $S^c$ | $(u,v) \in S$ | $(u,v) \in S^c$ |
| Chain(500,20) | 0.915 | 0.975 | 0.937 | 0.935 | 0.868 | 0.847 | 0.860 | 0.850 |
| Chain(500,50) | 0.917 | 0.987 | 0.927 | 0.988 | 0.847 | 0.831 | 0.846 | 0.816 |
| Chain(500,100) | 0.926 | 0.987 | 0.917 | 0.963 | 0.937 | 0.919 | 0.949 | 0.935 |
| NN(500,20) | 0.917 | 0.981 | 0.911 | 0.985 | 0.889 | 0.871 | 0.934 | 0.904 |
| NN(500,50) | 0.922 | 0.982 | 0.926 | 0.998 | 0.894 | 0.880 | 0.885 | 0.864 |
| NN(500,100) | 0.931 | 0.987 | 0.932 | 0.987 | 0.961 | 0.951 | 0.967 | 0.977 |
| Erdös-Rényi(500,20) | 0.954 | 0.982 | 0.935 | 0.981 | 1.268 | 1.283 | 1.292 | 1.238 |
| Erdös-Rényi(500,50) | 0.953 | 0.988 | 0.926 | 0.995 | 1.234 | 1.246 | 1.202 | 1.204 |
| Erdös-Rényi(500,100) | 0.950 | 0.993 | 0.921 | 0.972 | 1.293 | 1.312 | 1.307 | 1.299 |
| Scale-free(500,20) | 0.954 | 0.979 | 0.937 | 0.963 | 1.641 | 1.706 | 1.581 | 1.620 |
| Scale-free(500,50) | 0.973 | 0.992 | 0.985 | 0.998 | 1.885 | 1.951 | 1.810 | 1.826 |
| Scale-free(500,100) | 0.973 | 0.992 | 0.992 | 0.963 | 2.200 | 2.288 | 1.904 | 2.066 |

Table 3: Confidence interval, for sin function precision

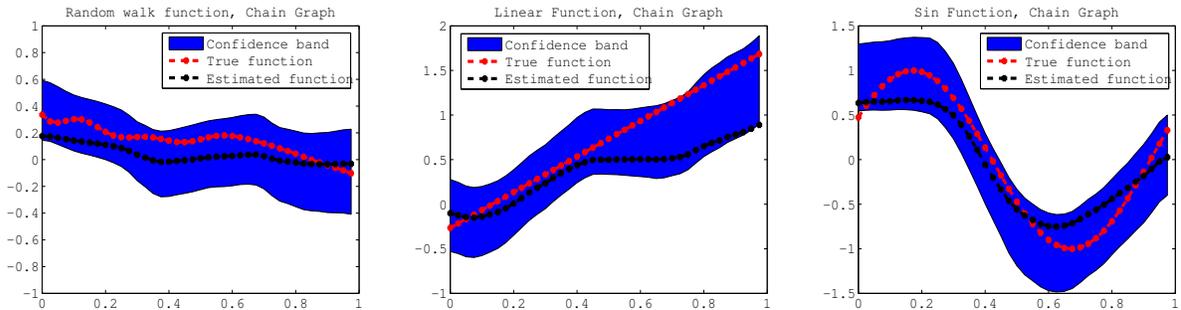

Figure 8: Illustration of the constructed confidence bands by the proposed procedure, on a $n = 500, p = 50$, chain graph problem.

# 6 Illustrative Applications to Real Data

In this section we apply our procedure on the stock price data, which is available in the "huge" package (Zhao et al., 2012). It contains 1258 closing prices of 452 stocks between January 1, 2003 and January 1, 2008. Instead of using the raw closing prices, we use logarithm of the ratio of prices at times $t$ and $t-1$. We further normalize the data for each stock to have mean 0 and variance 1. The 452 stocks are categorized into 10 Global Industry Classification Standard (GICS) sectors, including Consumer Discretionary, Consumer Staples, Energy, Financials, Health Care, Industrials, Information Technology, Telecommunications Services, Materials, and Utilities, each sector has 6 to 70 stocks.

We apply the glasso and our procedure to estimate associations between the stocks. In the estimated graph each node corresponds to a stock. An edge is put between two stocks if the corresponding element in the estimated inverse covariance matrix is not zero. When applying our procedure we consider the following as index variables: i) date, normalized to $[0, 1]$, ii) the Crude Oil price[1] (WTI - Cushing,Oklahoma), and iii) USD/EUR exchange rate[2]. Figure 9 shows the estimated graphs with colors representing different sectors. We can see that the graphs learned by $CCS_{\ell_1/\ell_2}$ give more clear cluster structure than the regular glasso graph.

We also consider a quantitative measure of the estimation performance. Since there is no true graph for this data, we use the Kullback-Leibler loss, as in Yuan and Lin (2007). First we divide the whole data set $[n]$ into $K$ folds: $\{D_1, D_2, ..., D_K\}$, then the $K$-fold cross-validation negative log-likelihood is

$$\sum_{k \in [K]} \sum_{i \in D_k} -\log \det(\widehat{\Omega}^{[n]/D_k}(\widetilde{z}^i)) + \text{tr}\left(\widehat{\Omega}^{[n]/D_k}(\widetilde{z}^i)\widetilde{x}^i \left(\widetilde{x}^i\right)^T\right),$$

where $\widehat{\Omega}^{[n]/D_k}(z)$ is learned from the training data set $[n]/D_k$ and $\{\widetilde{x}^i, \widetilde{z}^i\}_{i \in D_k}$ is a test set. For the regular glasso procedure, since it only estimate a single precision matrix, the above defined loss reduces to

$$-\sum_{k \in [K]} |D_k| \log \det(\widehat{\Omega}^{[n]/D_k}) + \text{tr}(\widehat{\Omega}^{[n]/D_k}\widehat{\Sigma}^{D_k})$$

where $\widehat{\Omega}^{[n]/D_k}$ is learned from the training data $[n]/D_k$, $\widehat{\Sigma}^{D_k}$ is the sample covariance matrix from the testing data $D_k$.

Table 4 reports 5-fold cross-validation score. We can see that for all index variable we considered here, our procedure improves over the regular glasso.

# 7 Discussion

We develop a new nonparametric method for high-dimensional conditional covariance selection. The method is developed under the assumption that the graph structure does not change with sample size. This assumption restricts the class of models and allows for better estimation when the sample size is small. Our theoretical results and simulation studies demonstrate this point. Furthermore, we develop point-wise confidence intervals for the individual elements in the model. These are useful for practitioners interested in making inferences and testing hypothesis about complex systems under investigation.

---

[1] http://www.eia.gov/dnav/pet/pet_pri_spt_s1_d.htm
[2] http://www.oanda.com/currency/historical-rates/

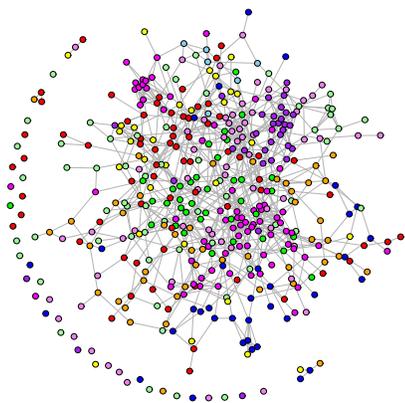
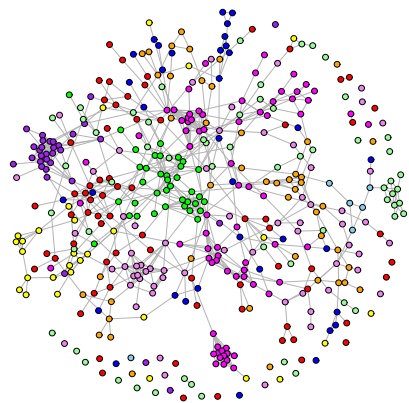
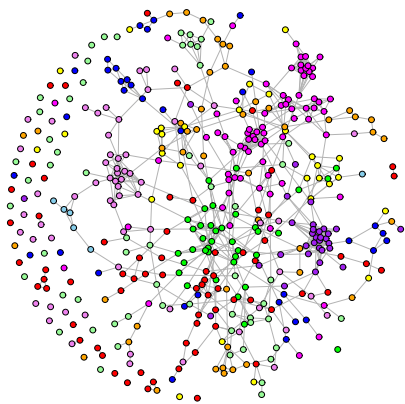
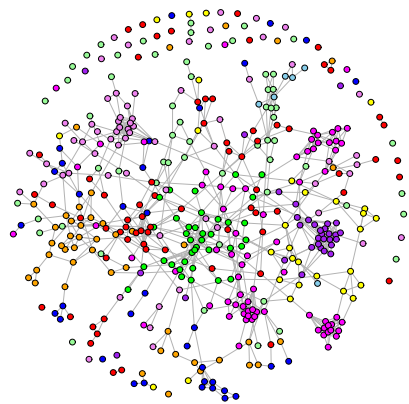

Figure 9: Graphical estimation on S&P 500 stock data.

| Procedure | 5-fold CV score |
|---|---|
| Glasso | $(1.5543 \pm 0.0302) \times 10^5$ |
| $\text{CCS}_{\ell_1/\ell_2}(\text{Z} = \text{time})$ | $(1.4612 \pm 0.0419) \times 10^5$ |
| $\text{CCS}_{\ell_1/\ell_2}(\text{Z} = \text{oil price})$ | $(1.5166 \pm 0.0434) \times 10^5$ |
| $\text{CCS}_{\ell_1/\ell_2}(\text{Z} = \text{USD/EUR X-rate})$ | $(1.4933 \pm 0.0409) \times 10^5$ |

Table 4: 5-fold cross-validation log negative likelihood

## A Proofs

### A.1 Proof of Theorem for Consistent Graph Recovery

In this section we provide a proof of Theorem 3. The proof is based on the primal-dual witness strategy described in Wainwright (2009) and Ravikumar et al. (2011). In the course of the proof we will construct a solution $\{\widehat{\Omega}(z^i)\}_{i \in [n]}$ to the optimization problem in (3) and show that it recovers the set $S$ correctly.

We will need the following restricted optimization problem

$$\min_{\substack{\{\Omega(z^i) \succ 0\}_{i \in [n]} \\ \{\Omega_{S^c}(z^i) = 0\}_{i \in [n]}}} \sum_{i \in [n]} \left( \text{tr}(\widehat{\Sigma}(z^i)\Omega(z^i)) - \log \det \Omega(z^i) \right) + \lambda \sum_{v \neq u} \sqrt{\sum_{i \in [n]} \Omega_{uv}(z^i)^2}. \quad (12)$$

Let $\{\widetilde{\Omega}(z^i)\}_{i \in [n]}$ be the unique solution to the restricted problem. Our strategy is to show the following

1. $\widehat{\Omega}_{uv}(z^i) = \widetilde{\Omega}_{uv}(z^i)$ and $\sum_{i \in [n]} \left(\widetilde{\Omega}_{uv}(z^i)\right)^2 \neq 0$ for $(u,v) \in S$ and $i \in [n]$;

2. $\widehat{\Omega}_{uv}(z^i) = 0$ for $(u,v) \in \bar{S}$ and $i \in [n]$,

hold with high probability. We will achieve these goals through the following steps

**Step 1:** Solve the restricted problem in (12) to obtain $\{\widetilde{\Omega}(z^i)\}_{i \in [n]}$.

**Step 2:** Compute the sub-differential $\{\widetilde{Z}(z^i)\}_{i \in [n]}$ of the penalty term $\sum_{v \neq u} \sqrt{\sum_{i \in [n]} (\Omega_{uv}(z^i))^2}$ at $\{\widetilde{\Omega}(z^i)\}_{i \in [n]}$, which can be obtained in a closed form as

$$\widetilde{Z}_{uv}(z^i) = \frac{\widetilde{\Omega}_{uv}(z^i)}{\sqrt{\sum_{i \in [n]} \left(\widetilde{\Omega}_{uv}(z^i)\right)^2}}.$$

**Step 3:** For each $(u,v) \in S^c$, set

$$\widetilde{Z}_{uv}(z^i) = \frac{1}{\lambda}\left( -\widehat{\Sigma}_{uv}(z^i) + \left(\widetilde{\Omega}(z)\right)^{-1}_{uv} \right).$$

**Step 4:** Verify that $\|\mathcal{H}^n(\widetilde{Z}_{S^c}(z))\|_\infty < n^{-1/2}$ and $\sum_{i \in [n]} \left(\widetilde{\Omega}_{uv}(z^i)\right)^2 \neq 0$.

In the first three steps, we construct the solution to the restricted optimization problem $\{\widetilde{\Omega}(z^i)\}_{i \in [n]}$ and the sub-differential $\{\widetilde{Z}(z^i)\}_{i \in [n]}$. In the forth step, we check whether the construction satisfies first order optimality conditions for the optimization program in (3) and that no elements of $S$ are excluded. The sub-differential of the penalty term in (3) evaluated at the solution $\{\widehat{\Omega}(z^i)\}_{i \in [n]}$ satisfies

$$\widehat{Z}_{uv}(z^i) = \begin{cases} 0 & \text{if } u = v, \\ \frac{\widehat{\Omega}_{uv}(z^i)}{\sqrt{\sum_{i \in [n]} \widehat{\Omega}_{uv}^2(z^i)}} & \text{if } u \neq v \text{ and } (u,v) \in \text{support}(\widehat{\Omega}(z)), \\ g_{uv}(z^i) \text{ s.t. } \sqrt{\sum_{i \in [n]} g_{uv}^2(z^i)} \leq 1 & \text{if } u \neq v \text{ and } (u,v) \notin \text{support}(\widehat{\Omega}(z)). \end{cases}$$

By showing that

$$\sqrt{\sum_{i \in [n]} \widetilde{Z}_{uv}^2(z^i)} < 1$$

holds with high probability, we establish that $\{\widetilde{\Omega}(z^i)\}_{i \in [n]}$ is the solution to the unrestricted optimization program (3) with high probability.

Lemma 8-11, given below, provide technical details. Let

$$\Delta(z) = \widetilde{\Omega}(z) - \Omega^*(z)$$

and

$$R(\Delta(z)) = \widetilde{\Omega}^{-1}(z) - \Omega^{*-1}(z) + \Omega^{*-1}(z)\Delta(z)\Omega^{*-1}(z).$$

Lemma 8 shows that with high probability

$$\left\|\mathcal{H}^n\left(\widehat{\Sigma}(z) - \Sigma^*(z)\right)\right\|_\infty \leq Cn^{-2/5}\sqrt{\log p}.$$

This is a main technical result that allows us to establish the desired rate of convergence. Using Lemma 8 we establish in Lemma 10 that

$$\|\mathcal{H}^n(\Delta(z))\|_\infty \lesssim \left\|\mathcal{H}^n\left(\widehat{\Sigma}(z) - \Sigma^*(z)\right)\right\|_\infty + n^{-1/2}\lambda.$$

Lemma 9 shows that

$$\|\mathcal{H}^n(R(\Delta(z)))\|_\infty \lesssim d\|\mathcal{H}^n(\Delta(z))\|_\infty$$

when $\|\mathcal{H}^n(\Delta(z))\|_\infty$ is sufficiently small. Lemma 11 obtains the strict dual feasibility and, finally, by applying Lemma 10 again we obtain that the procedure can identify the true support $S$ if

$$\sqrt{\frac{\sum_{i \in n} \Omega_{uv}^{*~2}(z^i)}{n}} \gtrsim n^{-2/5}\sqrt{\log p}, \qquad \forall (u,v) \in S.$$

**Lemma 8.** *Suppose that assumtions A1-A4 hold. If the bandwidth parameter satisfies $h \asymp n^{-1/5}$, then there exists a constant $C$, which depends only on $C_K$, $C_\Sigma$, and $\Lambda_\infty$, such that*

$$\|\mathcal{H}^n(\widehat{\Sigma}(z) - \Sigma^*(z))\|_\infty \leq C n^{-2/5} \sqrt{\log p}$$

*with probability at least $1 - \mathcal{O}(p^{-1})$.*

*Proof.* Lemma follows from the bound

$$\|\mathcal{H}^n(A(z))\|_\infty \leq \sup_{z \in [0,1]} \|A(z)\|_\infty$$

once we find a bound for $\sup_z \|\widehat{\Sigma}(z) - \Sigma^*(z)\|_\infty$.

Consider the following decomposition

$$\sup_z \|\widehat{\Sigma}(z) - \Sigma^*(z)\|_\infty \leq \sup_z \|\widehat{\Sigma}(z) - E\widehat{\Sigma}(z)\|_\infty + \sup_z \|E\widehat{\Sigma}(z) - \Sigma^*(z)\|_\infty.$$

For the bias term, $\sup_z \|E\widehat{\Sigma}(z) - \Sigma^*(z)\|_\infty$, we have

$$E\left[\widehat{\Sigma}_{uv}(z)\right] = E_Z\left[E_{X|Z}\left[\frac{\sum_{i=1}^n K_h(z^i - z) x_u^i x_v^i}{\sum_{i=1}^n K_H(z^i - z)}\right]\right]$$

$$= E_Z\left[\frac{\sum_{i=1}^n K_h(z^i - z) \Sigma^*_{uv}(z^i)}{\sum_{i=1}^n K_h(z^i - z)}\right]$$

$$= \frac{E_Z\left[\sum_{i=1}^n K_h(z^i - z) \Sigma^*_{uv}(z^i)\right]}{E_Z\left[\sum_{i=1}^n K_h(z^i - z)\right]} + \mathcal{O}\left(\frac{1}{nh}\right)$$

where the last step follows from Pagan and Ullah (1999, Page 101, (3.56)). Since

$$E_Z\left[K_h(z^i - z) \Sigma^*_{uv}(z^i)\right] = \frac{1}{h} \int K\left(\frac{z^i - z}{h}\right) \Sigma^*_{uv}(z^i) f(z^i) dz^i$$

$$= \int_{-1}^{1} K(a) \Sigma^*_{uv}(z + ah) f(z + ah) da$$

$$= \Sigma^*_{uv}(z) f(z) + \mathcal{O}(h^2)$$

and, similarly,

$$E_Z\left[K_h(z^i - z)\right] = f(z) + \mathcal{O}(h^2),$$

we obtain that

$$E\left[\widehat{\Sigma}_{uv}(z)\right] = \Sigma^*_{uv}(z) + \mathcal{O}(h^2) + \mathcal{O}\left(\frac{1}{nh}\right)$$

for any $u, v \in [p]$ and $z \in [0, 1]$. Therefore,

$$\sup_z \left\|E\left[\widehat{\Sigma}_{uv}(z)\right] - \Sigma^*_{uv}(z)\right\|_\infty = \mathcal{O}(h^2) + \mathcal{O}\left(\frac{1}{nh}\right). \tag{13}$$

For the variance term, $\sup_z \|\widehat{\Sigma}(z) - E\widehat{\Sigma}(z)\|_\infty$, we divide the interval $[0, 1]$ into $K$ segments $\{[(k-1)/K, k/K] \mid k \in [K]\}$, where $K$ will be chosen later. We have

$$
\begin{aligned}
\sup_z \|\widehat{\Sigma}(z) - E\widehat{\Sigma}(z)\|_\infty &= \max_{k \in [K]} \sup_{z \in \left[\frac{k-1}{K}, \frac{k}{K}\right]} \|\widehat{\Sigma}(z) - E\widehat{\Sigma}(z)\|_\infty \\
&\leq \max_{k \in [K]} \left\| \widehat{\Sigma}\left(\frac{2k-1}{2K}\right) - E\widehat{\Sigma}\left(\frac{2k-1}{2K}\right) \right\|_\infty \\
&\quad + \max_{k \in [K]} \sup_{z \in \left[\frac{k-1}{K}, \frac{k}{K}\right]} \left\| \widehat{\Sigma}(z) - E\widehat{\Sigma}(z) - \left( \widehat{\Sigma}\left(\frac{2k-1}{2K}\right) - E\widehat{\Sigma}\left(\frac{2k-1}{2K}\right) \right) \right\|_\infty.
\end{aligned}
\tag{14}
$$

We start with the first term in (14). We will use some standard results for the kernel density estimator

$$\widehat{f}(z) = \frac{1}{nh} \sum_{i \in [n]} K\left( \frac{z^i - z}{h} \right).$$

In particular, we have that

$$\sup_z \left| E\left[\widehat{f}(z)\right] - f(z) \right| = \mathcal{O}(h^2)$$

and

$$\sup_z \left| \widehat{f}(z) - E\left[\widehat{f}(z)\right] \right| = \mathcal{O}_P\left( \sqrt{\frac{\log(1/h)}{nh}} \right).$$

See, for example, Pagan and Ullah (1999) for the first result and Silverman (1978) for the second one. From these results, we have that

$$\sup_z \frac{\widehat{f}(z)}{f(z)} = 1 + \mathcal{O}(h^2) + \mathcal{O}_P\left( \sqrt{\frac{\log(1/h)}{nh}} \right). \tag{15}$$

Thus

$$
\begin{aligned}
\widehat{\Sigma}_{uv}\left(\frac{2k-1}{2K}\right) &= \frac{n^{-1} \sum_i K_h\left(z^i - \frac{2k-1}{2K}\right) x_u^i x_v^i}{n^{-1} \sum_i K_h\left(z^i - \frac{2k-1}{2K}\right)} \\
&= \frac{1}{nf\left(\frac{2k-1}{2K}\right)} \sum_i K_h\left(z^i - \frac{2k-1}{2K}\right) x_u^i x_v^i + \mathcal{O}(h^2) + \mathcal{O}_P\left( \sqrt{\frac{\log(1/h)}{nh}} \right).
\end{aligned}
$$

For a fixed $k \in [K]$, we have

$$
\begin{aligned}
&\left\| \widehat{\Sigma}\left(\frac{2k-1}{2K}\right) - E\left[\widehat{\Sigma}\left(\frac{2k-1}{2K}\right)\right] \right\|_\infty \\
&= \max_{u,v} \left| \widehat{\Sigma}_{uv}\left(\frac{2k-1}{2K}\right) - E\left[\widehat{\Sigma}_{uv}\left(\frac{2k-1}{2K}\right)\right] \right| \\
&= \max_{u,v} \frac{1}{nf\left(\frac{2k-1}{2K}\right)} \left| \sum_i K_h\left(z^i - \frac{2k-1}{2K}\right) x_u^i x_v^i - E\left[K_h\left(z^i - \frac{2k-1}{2K}\right) x_u^i x_v^i\right] \right| \\
&\quad + \mathcal{O}(h^2) + \mathcal{O}_P\left( \sqrt{\frac{\log(1/h)}{nh}} \right).
\end{aligned}
$$

Conditionally of $z^i$, we have

$$\|x_u^i x_v^i\|_{\Psi_1} \leq 2\|e_a^T x_i\|_{\Psi_2}\|e_b^T x_i\|_{\Psi_2} = 2\Sigma_{uu}^{1/2}(z^i)\Sigma_{vv}^{1/2}(z^i)$$

and

$$\|x_u^i x_v^i - E\left[x_u^i x_v^i \mid z^i\right]\|_{\Psi_1} \leq 2\|x_u^i x_v^i\|_{\Psi_1} \leq 4\Lambda_\infty.$$

Conditionally on $\{z^i\}_{i \in [n]}$, Proposition 5.16 in Vershynin (2012) gives us that

$$\left|\sum_{i \in [n]} \left[K\left(\frac{z^i - z}{h}\right)(x_u^i x_v^i - E[x_u^i x_v^i \mid z^i])\right]\right|$$
$$\lesssim M_\sigma \left(\sqrt{\sum_{i \in [n]} K\left(\frac{z^i - z}{h}\right)^2 \log(2/\delta)} \bigvee \left(\max_{i \in [n]} K\left(\frac{z^i - z}{h}\right)\right) \log(2/\delta)\right)$$

with probability at least $1 - \delta$. We have

$$E\left[K\left(\frac{z^i - z}{h}\right)^2\right] = \int K^2\left(\frac{x - z}{h}\right) f_Z(x) dx = \mathcal{O}(h)$$

and, similarly,

$$\mathrm{var}\left[K\left(\frac{z^i - z}{h}\right)^2 - E\left[K\left(\frac{z^i - z}{h}\right)^2\right]\right] \leq E\left[K\left(\frac{z^i - z}{h}\right)^4\right] = \mathcal{O}(h).$$

Since

$$\left|K\left(\frac{z^i - z}{h}\right)^2 - E\left[K\left(\frac{z^i - z}{h}\right)\right]\right| \leq 2C_K^2 \text{ a.s.,}$$

we can use Bernstein's inequality to obtain

$$\sum_{i \in [n]} K\left(\frac{z^i - z}{h}\right)^2 \lesssim nh + C\sqrt{nh \log(1/\delta)}$$

with probability at least $1 - \delta$. Therefore, for a fixed $u, v$ and $k$, we obtain

$$\frac{1}{nf\left(\frac{2k-1}{2K}\right)}\left|\sum_i K_h\left(z^i - \frac{2k-1}{2K}\right) x_u^i x_v^i - E\left[K_h\left(z^i - \frac{2k-1}{2K}\right) x_u^i x_v^i\right]\right|$$
$$\lesssim \sqrt{\frac{\log(1/\delta)}{nh}}$$

with probability $1 - 2\delta$. An application of the union bound gives us

$$\mathrm{pr}\left(\max_{k \in [K]} \left\|\widehat{\Sigma}\left(\frac{2k-1}{2K}\right) - E\left[\widehat{\Sigma}\left(\frac{2k-1}{2K}\right)\right]\right\|_\infty \geq \epsilon\right) \leq C_1 \exp\left(-C_2\left(nh\epsilon^2 + \log(pK)\right)\right). \tag{16}$$

For the second term in (14), we have

$$\left\|\widehat{\Sigma}(z) - E\widehat{\Sigma}(z) - \left(\widehat{\Sigma}\left(\frac{2k-1}{2K}\right) - E\widehat{\Sigma}\left(\frac{2k-1}{2K}\right)\right)\right\|_\infty \tag{17}$$
$$\leq \left\|E\widehat{\Sigma}(z) - E\widehat{\Sigma}\left(\frac{2k-1}{2K}\right)\right\|_\infty + \left\|\widehat{\Sigma}(z) - \widehat{\Sigma}\left(\frac{2k-1}{2K}\right)\right\|_\infty.$$

We further have

$$\sup_{z\in[\frac{k-1}{K},\frac{k}{K}]} \left\|E\widehat{\Sigma}(z) - E\widehat{\Sigma}\left(\frac{2k-1}{2K}\right)\right\|_\infty$$
$$\leq \sup_{z\in[\frac{k-1}{K},\frac{k}{K}]} \left\|E\widehat{\Sigma}(z) - \Sigma^*(z)\right\|_\infty + \sup_{z\in[\frac{k-1}{K},\frac{k}{K}]} \left\|\Sigma^*(z) - \Sigma^*\left(\frac{2k-1}{2K}\right)\right\|_\infty$$
$$+ \left\|\Sigma^*\left(\frac{2k-1}{2K}\right) - E\widehat{\Sigma}\left(\frac{2k-1}{2K}\right)\right\|_\infty$$
$$= \mathcal{O}(h^2) + \mathcal{O}\left(\frac{1}{nh}\right) + \mathcal{O}(K^{-2}).$$

using (13) and the smoothness of $\Sigma^*(z)$. We also have

$$\sup_{z\in[\frac{k-1}{K},\frac{k}{K}]} \left\|\widehat{\Sigma}(z) - \widehat{\Sigma}\left(\frac{2k-1}{2K}\right)\right\|_\infty$$
$$= \max_{u,v} \sup_{z\in[\frac{k-1}{K},\frac{k}{K}]} \left|\frac{n^{-1}\sum_i K_h(z^i - z) x_u^i x_v^i}{f(z)} - \frac{n^{-1}\sum_i K_h\left(z^i - \frac{2k-1}{2K}\right) x_u^i x_v^i}{f(\frac{2k-1}{2K})}\right|$$
$$+ \mathcal{O}(h^2) + \mathcal{O}_P\left(\sqrt{\frac{\log(1/h)}{nh}}\right)$$

using the property of $\widehat{f}(z)$ in (15). Using smoothness of $f(z)$,

$$\sup_{z\in[\frac{k-1}{K},\frac{k}{K}]} \left|(f(z))^{-1} - \left(f\left(\frac{2k-1}{2K}\right)\right)^{-1}\right| = \mathcal{O}(K^{-2})$$

and
$$\sup_{z\in[\frac{k-1}{K},\frac{k}{K}]} \left\| \widehat{\Sigma}(z) - \widehat{\Sigma}\left(\frac{2k-1}{2K}\right) \right\|_{\infty}$$
$$= \max_{u,v} \sup_{z\in[\frac{k-1}{K},\frac{k}{K}]} \frac{1}{nf(z)} \left| \sum_i K_h\left(z^i - z\right) x_u^i x_v^i - K_h\left(z^i - \frac{2k-1}{2K}\right) x_u^i x_v^i \right|$$
$$+ \mathcal{O}_P(h^2) + \mathcal{O}_P\left(\sqrt{\frac{\log(1/h)}{nh}}\right) + \mathcal{O}_P(K^{-2})$$
$$\leq \max_{u,v} \sup_{z\in[\frac{k-1}{K},\frac{k}{K}]} \left| \frac{1}{nhf(z)} \sum_{|z^i-z|\leq h, |z^i-\frac{2k-1}{2K}|\leq h} \frac{C_L}{Kh} x_u^i x_v^i \right|$$
$$+ \max_{u,v} \sup_{z\in[\frac{k-1}{K},\frac{k}{K}]} \left| \frac{1}{nhf(z)} \sum_{|z^i-z|> h, |z^i-\frac{2k-1}{2K}|\leq h} C_K x_u^i x_v^i \right|$$
$$+ \max_{u,v} \sup_{z\in[\frac{k-1}{K},\frac{k}{K}]} \left| \frac{1}{nhf(z)} \sum_{|z^i-z|\leq h, |z^i-\frac{2k-1}{2K}|> h} C_K x_u^i x_v^i \right|$$
$$+ \mathcal{O}(h^2) + \mathcal{O}_P\left(\sqrt{\frac{\log(1/h)}{nh}}\right) + \mathcal{O}(K^{-2})$$
$$= \mathcal{O}(h^2) + \mathcal{O}_P\left(\sqrt{\frac{\log(1/h)}{nh}}\right) + \mathcal{O}(K^{-2}) + \mathcal{O}_P\left((Kh)^{-1}\right).$$

Choosing $K \asymp n$ and inserting back into (17), we obtain
$$\max_{k\in[K]} \sup_{z\in[\frac{k-1}{K},\frac{k}{K}]} \left\| \widehat{\Sigma}(z) - E\widehat{\Sigma}(z) - \left(\widehat{\Sigma}\left(\frac{2k-1}{2K}\right) - E\widehat{\Sigma}\left(\frac{2k-1}{2K}\right)\right) \right\|_{\infty}$$
$$= \mathcal{O}(h^2) + \mathcal{O}_P\left(\sqrt{\frac{\log(1/h)}{nh}}\right) + \mathcal{O}_P\left((nh)^{-1}\right). \quad (18)$$

Combining (13), (14), (16), and (18), together with $K \asymp n$ and $h \asymp n^{-1/5}$, we get
$$\text{pr}(\sup_z \|\widehat{\Sigma}(z) - \Sigma^*(z)\|_\infty \geq \epsilon) \leq C_1 \exp(-C_2 n^{4/5} \epsilon + \log p).$$

Choosing $\epsilon \asymp n^{-2/5}\sqrt{\log p}$ completes the proof. □

**Lemma 9.** *Under conditions of Theorem 3, if*
$$\|\mathcal{H}^n(\Delta(z))\|_\infty \leq \frac{1}{3C_\infty d},$$
*then*
$$\|\mathcal{H}^n(R(\Delta(z)))\|_\infty \leq \frac{3d}{2} C_\infty^3 \|\mathcal{H}^n(\Delta(z))\|_\infty^2.$$

*Proof.* The remainder term can be re-written as

$$R(\Delta(z)) = (\Omega^*(z) + \Delta(z))^{-1} - \Omega^*(z) + \Omega^{*-1}(z)\Delta(z)\Omega^{*-1}(z).$$

We have

$$\begin{aligned}\|\mathcal{H}^n(\Omega^{*-1}(z)\Delta(z))\|_{\infty,\infty} &\leq \|\mathcal{H}^n(\Omega^{*-1}(z))\|_{\infty,\infty}\|\mathcal{H}^n(\Delta(z))\|_{\infty,\infty} \\ &\leq \|\mathcal{H}^n(\Omega^{*-1}(z))\|_{\infty,\infty} d \|\mathcal{H}^n(\Delta(z))\|_{\infty} \\ &\leq \frac{1}{3}.\end{aligned}$$

where the first step uses facts in Lemma 1, and the second step is based on the fact that $\Delta(z)$ has at most $d$ non-zero entries per row/column. Based on convergent matrix expansion

$$\begin{aligned}(\Omega^*(z) + \Delta(z))^{-1} &= \sum_{k=0}^{\infty}(-1)^k(\Omega^{*-1}(z)\Delta(z))^k\Omega^{*-1}(z) \\ &= \Omega^{*-1}(z) - \Omega^{*-1}(z)\Delta(z)\Omega^{*-1}(z) + \Omega^{*-1}(z)\Delta(z)\Omega^{*-1}(z)\Delta(z)J(z)\Omega^{*-1}(z)\end{aligned}$$

where $J(z) = \sum_{k=0}^{\infty}(-1)^k(\Omega^{*-1}(z)\Delta(z))^k$. Thus

$$\begin{aligned}\|\mathcal{H}^n(R(\Delta(z)))\|_{\infty} &= \|\mathcal{H}^n(\Omega^{*-1}(z)\Delta(z)\Omega^{*-1}(z)\Delta(z)J(z)\Omega^{*-1}(z))\|_{\infty} \\ &\leq \|\mathcal{H}^n(\Omega^{*-1}(z)\Delta(z))\|_{\infty,\infty}\|\mathcal{H}^n(\Omega^{*-1}(z)\Delta(z)J(z)\Omega^{*-1}(z))\|_{\infty} \\ &\leq (\|\mathcal{H}^n(\Omega^{*-1}(z))\|_{\infty,\infty})^3\|\mathcal{H}^n(\Delta(z))\|_{\infty}\|\mathcal{H}^n(J(z))\|_{\infty,\infty}\|\mathcal{H}^n(\Delta(z))\|_{\infty,\infty} \\ &\leq dC_{\infty}^3(\|\mathcal{H}^n(\Delta(z))\|_{\infty})^2\|\mathcal{H}^n(J(z))\|_{\infty,\infty}\end{aligned}$$

Where above steps use Lemma 1, the fact that $\Delta(z)$ has at most $d$ non-zero entries per row/column, and A2. Furthermore, we have a bound on $\|\mathcal{H}^n(J(z))\|_{\infty,\infty}$ as

$$\begin{aligned}\|\mathcal{H}^n(J(z))\|_{\infty,\infty} &= \|\mathcal{H}^n(\sum_{k=0}^{\infty}(-1)^k(\Omega^{*-1}(z)\Delta(z))^k)\|_{\infty,\infty} &&\text{(Definition of } J(z)\text{)} \\ &\leq \sum_{k=0}^{\infty}\|\mathcal{H}^n((\Omega^{*-1}(z)\Delta(z))^k)\|_{\infty,\infty} &&\text{(Triangular inequality)} \\ &\leq \sum_{k=0}^{\infty}(\|\mathcal{H}^n(\Omega^{*-1}(z)\Delta(z))\|_{\infty,\infty})^k &&\text{(Lemma 1)} \\ &\leq \frac{1}{1 - \|\mathcal{H}^n(\Omega^{*-1}(z)\Delta(z))\|_{\infty,\infty}} &&(\|\mathcal{H}^n(\Omega^{*-1}(z)\Delta(z))\|_{\infty,\infty} \leq \frac{1}{3}) \\ &\leq \frac{3}{2}.\end{aligned}$$

Combining the results

$$\|\mathcal{H}^n(R(\Delta(z)))\|_{\infty} \leq \frac{3d}{2}C_{\infty}^3\|\mathcal{H}^n(\Delta(z))\|_{\infty}^2$$

as claimed. □

**Lemma 10.** *Under conditions of Theorem 3, if*

$$r := 2C_{\mathcal{I}}(\|\mathcal{H}^n(\widehat{\Sigma}(z) - \Sigma^*(z))\|_\infty + n^{-1/2}\lambda) \leq \min(\frac{1}{3C_\infty d}, \frac{1}{3C_{\mathcal{I}}C_\infty^3 d}),$$

*then*

$$\|\mathcal{H}^n(\Delta(z))\|_\infty \leq r.$$

*Proof.* Define the gradient mapping function as

$$G(\Omega_S(z)) = -(\Omega^{-1}(z))_S + \widehat{\Sigma}_S(z) + \lambda \widetilde{Z}_S(z).$$

Construct the following function $F : \mathbb{R}^{|S| \times n} \to \mathbb{R}^{|S| \times n}$ via

$$F(\mathcal{H}^n(\widehat{\Delta}_S(z))) := -\mathcal{H}^n((\mathcal{I}_{SS}(z))^{-1}G(\Omega_S^*(z) + \widehat{\Delta}_S(z))) + \mathcal{H}^n(\widehat{\Delta}_S(z))$$

By construction, $F(\mathcal{H}^n(\Delta_S(z))) = \mathcal{H}^n(\Delta_S(z))$ since $G(\widetilde{\Omega}_S(z)) = (\Omega_S^*(z) + \Delta_S(z)) = 0$ based on the zero-gradient condition. Define the ball $\mathbb{B}(r)$ as

$$\mathbb{B}(r) := \{A : \|A\|_\infty \leq r\}.$$

The next thing we need to shows is $F(\mathbb{B}(r)) \subseteq \mathbb{B}(r)$, thus by fix point theorem, we can conclude $\|\mathcal{H}^n(\Delta(z))\|_\infty \leq r$. For any $\mathcal{H}(\widehat{\Delta}_S(z)) \in \mathbb{B}(r)$, first decompose $F(\mathcal{H}^n(\widehat{\Delta}_S(z)))$ as

$$F(\mathcal{H}^n(\widehat{\Delta}_S(z))) = \mathcal{H}^n((\mathcal{I}_{SS}(z))^{-1}R(\widehat{\Delta}(z))_S) - \mathcal{H}^n((\mathcal{I}_{SS}(z))^{-1}(\widehat{\Sigma}_S(z) - \Sigma_S^*(z) + \lambda \widetilde{Z}_S(z))), \tag{19}$$

and bound the two terms of above equation separately: for the first term, we have

$$\|\mathcal{H}^n((\mathcal{I}_{SS}(z))^{-1}R(\widehat{\Delta}(z))_S)\|_\infty \leq \|\mathcal{H}^n((\mathcal{I}_{SS}(z))^{-1})\|_\infty \|\mathcal{H}^n(R(\Delta(z))_S)\|_\infty$$
$$\leq C_{\mathcal{I}}\|\mathcal{H}^n(R(\widehat{\Delta}(z)))\|_\infty.$$

Since we assume $\mathcal{H}^n(\widehat{\Delta}_S(z)) \in \mathbb{B}(r)$, applying Lemma 9, we get

$$\begin{aligned}\|\mathcal{H}^n((\mathcal{I}_{SS}(z))^{-1}R(\widehat{\Delta}(z))_S)\|_\infty &\leq \|\mathcal{H}^n((\mathcal{I}_{SS}(z))^{-1})\|_{\infty,\infty}\|\mathcal{H}^n(R(\widehat{\Delta}(z)))\|_\infty\\ &\leq C_{\mathcal{I}}\|\mathcal{H}^n(R(\widehat{\Delta}(z)))\|_\infty\\ &\leq C_{\mathcal{I}}\frac{3d}{2}C_\infty^3\|\mathcal{H}^n(\widehat{\Delta}(z))\|_\infty^2\\ &\leq C_{\mathcal{I}}\frac{3d}{2}C_\infty^3 r^2\\ &\leq C_{\mathcal{I}}\frac{3d}{2}C_\infty^3\frac{1}{3C_{\mathcal{I}}C_\infty^3 d}r\\ &\leq \frac{r}{2}.\end{aligned} \tag{20}$$

For the second term, we have

$$\begin{aligned}
\|\mathcal{H}^n((\mathcal{I}_{SS}(z))^{-1}(\widehat{\Sigma}_S(z) - \Sigma_S^*(z) + \lambda \widetilde{Z}_S(z)))\|_\infty \\
\leq C_\mathcal{I}\|\mathcal{H}^n(\widehat{\Sigma}(z) - \Sigma^*(z))\|_\infty + C_\mathcal{I}\lambda\|\mathcal{H}^n(\widetilde{Z}_S(z))\|_\infty \\
\leq C_\mathcal{I}(\|\mathcal{H}^n(\widehat{\Sigma}(z) - \Sigma^*(z))\|_\infty + n^{-1/2}\lambda) \\
= \frac{r}{2}.
\end{aligned} \quad (21)$$

Substituting (20) and (21) into (19) we get $F(\mathcal{H}^n(\widehat{\Delta}_S(z))) \in \mathbb{B}(r)$, which concludes the proof. $\square$

**Lemma 11.** *Under conditions of Theorem 3, if*

$$\|\mathcal{H}^n(\widehat{\Sigma}(z) - \Sigma^*(z))\|_\infty \leq \frac{n^{-1/2}\alpha\lambda}{8} \quad \text{and} \quad \|\mathcal{H}^n(R(\Delta(z)))\|_\infty \leq \frac{n^{-1/2}\alpha\lambda}{8},$$

*then $\widetilde{Z}_{S^c}$ constructed in Step 3 of the primal-dual witness process satisfies $\|\mathcal{H}^n(\widetilde{Z}_{S^c}(z))\|_\infty < n^{-1/2}$.*

*Proof.* Based on the construction $\forall i \in [n], \widetilde{Z}_{uv}(z^i) = \frac{1}{\lambda}(-\widehat{\Sigma}(z^i)_{uv} + \widetilde{\Omega}_{uv}^{-1}(z^i))$. Substituting with the definition of $\Delta(z)$ and $R(\Delta(z))$, we have $\forall i \in [n]$

$$\Omega^*(z^i)^{-1}\Delta(z^i)\Omega^*(z^i)^{-1} + \widehat{\Sigma}(z^i) - \Omega^*(z^i)^{-1} - R(\Delta(z^i)) + \lambda \widetilde{Z}(z^i) = 0 \quad (22)$$

Since

$$\text{vec}(\Omega^*(z^i)^{-1}\Delta(z^i)\Omega^*(z^i)^{-1}) = (\Omega^*(z^i)^{-1} \otimes \Omega^*(z^i)^{-1})\bar{\Delta}(z^i) = \mathcal{I}(z^i)\bar{\Delta}(z^i)$$

thus we can re-write Equation (22) as two blocks of linear equations

$$\mathcal{I}_{SS}(z^i)\bar{\Delta}_S(z^i) + \bar{\widehat{\Sigma}}_S(z^i) - \bar{\Omega}^*(z^i)_S^{-1} - \bar{R}(\Delta(z^i))_S + \lambda \bar{\widetilde{Z}}_S(z^i) = 0 \quad (23)$$

$$\mathcal{I}_{S^cS}(z^i)\bar{\Delta}_S(z^i) + \bar{\widehat{\Sigma}}_{S^c}(z^i) - \bar{\Omega}^*(z^i)_{S^c}^{-1} - \bar{R}(\Delta(z^i))_{S^c} + \lambda \bar{\widetilde{Z}}_{S^c}(z^i) = 0 \quad (24)$$

Solving $\bar{\Delta}_S(z^i)$ using Equation (23), we get

$$\bar{\Delta}_S(z^i) = \mathcal{I}_{SS}^{-1}(z^i)[\bar{R}(\Delta(z^i))_S - \bar{\widehat{\Sigma}}_S(z^i) + \bar{\Omega}^*(z^i)_S^{-1} - \lambda \bar{\widetilde{Z}}_S(z^i)]$$

substituting above to Equation (24), we can solve $\bar{\widetilde{Z}}_{S^c}(z^i)$ as

$$\begin{aligned}
\bar{\widetilde{Z}}_{S^c}(z^i) &= -\frac{1}{\lambda}\mathcal{I}_{S^cS}(z^i)\bar{\Delta}_S(z^i) + \frac{1}{\lambda}\bar{R}(\Delta(z^i))_{S^c} - \frac{1}{\lambda}(\bar{\widehat{\Sigma}}_{S^c}(z^i) - \bar{\Omega}^*(z^i)_{S^c}^{-1}) \\
&= -\frac{1}{\lambda}\mathcal{I}_{S^cS}(z^i)\mathcal{I}_{SS}^{-1}(z^i)[\bar{R}(\Delta(z^i))_S - \bar{\widehat{\Sigma}}_S(z^i) + \bar{\Omega}^*(z^i)_S^{-1} - \lambda \bar{\widetilde{Z}}_S(z^i)] \\
&\quad + \frac{1}{\lambda}\bar{R}(\Delta(z^i))_{S^c} - \frac{1}{\lambda}(\bar{\widehat{\Sigma}}_{S^c}(z^i) - \bar{\Omega}^*(z^i)_{S^c}^{-1})
\end{aligned}$$

Thus

$$\|\mathcal{H}^n(\widetilde{Z}_{S^c}(z))\|_\infty \leq \frac{1}{\lambda}\|\mathcal{H}^n(\mathcal{I}_{S^cS}(z)\mathcal{I}_{SS}^{-1}(z)[R(\Delta(z))_S - \widehat{\Sigma}_S(z) + \Omega^*(z)_S^{-1} - \lambda\widetilde{Z}_S(z)])\|_\infty$$
$$+ \frac{1}{\lambda}\|R(\Delta(z))_{S^c}\|_\infty + \frac{1}{\lambda}\|\mathcal{H}^n(\widehat{\Sigma}_{S^c}(z^i) - \Omega^*(z)_{S^c}^{-1})\|_\infty$$
$$\leq \frac{1}{\lambda}\|\mathcal{H}^n(\mathcal{I}_{S^cS}(z)(\mathcal{I}_{SS}(z))^{-1})\|_{\infty,\infty}(\|\mathcal{H}^n(\widehat{\Sigma}(z) - \Sigma^*(z))\|_\infty + \|\mathcal{H}^n(R(\Delta(z)))\|_\infty)$$
$$+ \|\mathcal{H}^n(\mathcal{I}_{S^cS}(z)(\mathcal{I}_{SS}(z))^{-1})\|_{\infty,\infty}\|\mathcal{H}^n(\widetilde{Z}_S(z))\|_\infty$$
$$+ \frac{1}{\lambda}\frac{n^{-1/2}\alpha\lambda}{8} + \frac{1}{\lambda}\frac{n^{-1/2}\alpha\lambda}{8}$$
$$\leq \frac{(1-\alpha)}{\lambda}\frac{n^{-1/2}\alpha\lambda}{4} + n^{-1/2}(1-\alpha) + \frac{n^{-1/2}\alpha}{4} \leq \frac{n^{-1/2}\alpha}{2} + n^{-1/2}(1-\alpha)$$
$$< n^{-1/2}.$$

$\square$

## A.2 Proof of Proposition 5

*Proof.* Recall that

$$\frac{1}{n}\sum_{i=1}^n \|\widehat{\Omega}(z^i) - \Omega^*(z^i)\|_F^2 = \|\mathcal{H}^n(\widehat{\Omega}(z) - \Omega^*(z))\|_F^2.$$

Under the assumption of Theorem 3, the pattern of non-zero elements is recovered correctly. Therefore, there are $|S| + p$ non-zero entries in the matrix $\mathcal{H}^n(\widehat{\Omega}(z) - \Omega^*(z))$. We have

$$\|\mathcal{H}^n(\widehat{\Omega}(z) - \Omega^*(z))\|_F \leq \sqrt{|S| + p}\|\mathcal{H}^n(\widehat{\Omega}(z) - \Omega^*(z))\|_\infty$$

and the proof follows from an application of Lemma 10 to the right hand side. $\square$

## A.3 Proof of Theorem for Construction of Confidence Intervals

We start the proof by stating the following relationship

$$\widehat{T}_{uv}(z) = \Omega^*_{uv}(z) - e_{uv}^T\left(\Omega^*(z) \otimes \Omega^*(z)\right)\text{vec}\left(\widehat{\Sigma}(z) - \Sigma^*(z)\right) + o_P(n^{-3/8})$$

which holds for all $z \in \{z^i : i \in [n]\}$ and $u, v \in [p]$. This relationship is established in Lemma 12 at the end of the section. Furthermore

$$n^{3/8}\left(\Omega^*(z) \otimes \Omega^*(z)\right)\text{vec}\left(\widehat{\Sigma}(z) - \Sigma^*(z)\right)$$
$$= n^{3/8}\left(\Omega^*(z) \otimes \Omega^*(z)\right)\text{vec}\left(\widehat{\Sigma}(z) - E\widehat{\Sigma}(z)\right) + n^{3/8}\left(\Omega^*(z) \otimes \Omega^*(z)\right)\text{vec}\left(E\widehat{\Sigma}(z) - \Sigma^*(z)\right)$$

From (13) we have $\|E\widehat{\Sigma}(z) - \Sigma^*(z)\|_\infty = \mathcal{O}(h^2) = \mathcal{O}(n^{-1/2})$. Combined with the fact that $\Omega^*(z)$ is sparse, we have

$$\left\|n^{3/8}\left(\Omega^*(z) \otimes \Omega^*(z)\right)\text{vec}\left(E\widehat{\Sigma}(z) - \Sigma^*(z)\right)\right\|_\infty = o(1).$$

Putting these results together, we have

$$n^{3/8}\left(\widehat{T}_{uv}(z) - \Omega^*_{uv}(z)\right) = -e^T_{uv}\left(\Omega^*(z) \otimes \Omega^*(z)\right) \text{vec}\left(\widehat{\Sigma}(z) - \Sigma^*(z)\right) + o_P(1).$$

Define the variance normalization function $V(z) : [0,1] \to \mathbb{R}^{p \times p}$ as

$$V_{uv}(z) = \sqrt{\frac{\widehat{f}(z)}{\left(\widehat{\Omega}^2_{uv}(z) + \widehat{\Omega}_{uu}(z)\widehat{\Omega}_{vv}(z)\right)\int_{-\infty}^{\infty} K^2(t)\mathrm{d}t}}.$$

We proceed to show that

$$n^{3/8} V_{uv}(z) e^T_{uv}\left(\Omega^*(z) \otimes \Omega^*(z)\right) \text{vec}\left(\widehat{\Sigma}(z) - E\widehat{\Sigma}(z)\right)$$

converges in distribution to $\mathcal{N}(0,1)$. The result will then follow from an application of Lemma 12 given below.

We first consider the elements of $n^{3/8}\left(\Omega^*(z) \otimes \Omega^*(z)\right)\text{vec}\left(\widehat{\Sigma}(z) - E\widehat{\Sigma}(z)\right)$ and show that they are asymptotically Normal with bounded variance. Fix $u, v$, and $z$, and define the random variable $J_{uv}(z)$ as

$$J_{uv}(z) = \sum_{i \in [n]} L^i_{uv}(z),$$

where

$$L^i_{uv}(z) = \left(\Omega^*(z)\left(w^i(z)x^i(x^i)^T - Ew^i(z)x^i(x^i)^T\right)\Omega^*(z)\right)_{uv}.$$

We have that $E\left[L^i_{uv}(z)\right] = 0$ and $\text{var}\left[L^i_{uv}(z)\right] < \infty$. Let

$$s_n^2 = \sum_i \text{var}\left[L^i_{uv}(z)\right] = n^{3/4}\text{var}\left[J_{uv}(z)\right].$$

In order to establish asymptotic Normality, we proceed to show that the Lindeberg condition is satisfied, that is, for all $\varepsilon > 0$

$$\lim_{n \to \infty} \frac{1}{s_n^2} \sum_i E\left[(L^i_{uv}(z))^2 \mathbb{I}(|L^i_{uv}(z)| > \varepsilon s_n)\right] = 0. \tag{25}$$

Since $L^i_{uv}(z)$ is a sub-exponential random variable, (25) is easily verified. We only need to show

$$\lim_{n \to \infty} E\left[(L^i_{uv}(z))^2 \mathbb{I}(|L^i_{uv}(z)| > \varepsilon s_n)\right] = 0.$$

Using Fubini's theorem, we have

$$E\left[(L^i_{uv}(z))^2 \mathbb{I}(|L^i_{uv}(z)| > \varepsilon s_n)\right] = 2\int_{\varepsilon s_n}^{\infty} t\,\text{pr}\left[L^i_{uv}(z) > t\right]dt + \varepsilon^2 s_n^2 \text{pr}\left[|L^i_{uv}(z)| > \varepsilon s_n\right].$$

Therefore, we will prove that

$$\lim_{n \to \infty} \varepsilon^2 s_n^2 \text{pr}\left[|L^i_{uv}(z)| > \varepsilon s_n\right] = 0$$

and
$$\lim_{n\to\infty} \int_{\varepsilon s_n}^{\infty} t\mathrm{pr}\left[L_{uv}^i(z) > t\right] dt = 0.$$

Since $|L_{uv}^i(z)|$ is a sub-exponential random variable, we have
$$\mathrm{pr}\left[|L_{uv}^i(z)| > \varepsilon s_n\right] \leq c_1 \exp(-c_2\varepsilon s_n)$$

for some constant $c_1, c_2$. Note that $\mathrm{var}\left[L_{uv}^i(z)\right]$ is both upper and lower bounded, thus $s_n \asymp n^{3/8}$. We have
$$\lim_{n\to\infty} \varepsilon^2 s_n^2 \mathrm{pr}\left[|L_{uv}^i(z)| > \varepsilon s_n\right] \leq \lim_{n\to\infty} \varepsilon^2 n^{3/4} c_1 \exp(-c_2\varepsilon n^{3/8}) = 0$$

and
$$\lim_{n\to\infty} \int_{\varepsilon s_n}^{\infty} u\mathrm{pr}\left[L_{uv}^i(z) > u\right] du \leq \lim_{n\to\infty} \int_{\varepsilon n^{3/8}}^{\infty} uc_1 \exp(-c_2\varepsilon n^{3/8}) du = 0.$$

This verifies the Lindenberg condition in (25) and we have that
$$\frac{J_{uv}(z)}{\sqrt{\mathrm{var}\left[J_{uv}(z)\right]}} \to_D \mathcal{N}(0,1).$$

The proof will be complete once we establish
$$nhV_{uv}^2(z)\,\mathrm{var}\left[J_{uv}(z)\right] \longrightarrow_P 1.$$

Note that for every $i \in [n]$ such that $K_h(z^i - z) \neq 0$ we have $\|\Omega^*(z) - \Omega^*(z^i)\|_\infty = \mathcal{O}(h^2)$ and $\|\Omega^*(z) - \Omega^*(z^i)\|_{\infty,\infty} = \mathcal{O}(h^2)$. Therefore,

$$\begin{aligned}
&\mathrm{var}\left[J_{uv}(z)\right] \\
&= \mathrm{var}\left[\sum_{i\in[n]} e_u^T \Omega^*(z)\left(w^i(z)x^i x^{iT} - E\left[w^i(z)x^i x^{iT}\right]\right)\Omega^*(z)e_v\right] \\
&= \frac{1}{(nhf(z))^2} \sum_{i\in[n]} E\left[\left(K(h^{-1}(z^i-z))(\Omega^*(z)x^i)_u(\Omega^*(z)x^i)_v\right.\right. \\
&\qquad\qquad\left.\left. - E\left[K(h^{-1}(z^i-z))(\Omega^*(z)x^i)_u(\Omega^*(z)x^i)_v\right]\right)^2\right] + o(1) \\
&= \frac{\Omega_{uu}^*(z)\Omega_{vv}^*(z) + \Omega_{uv}^{*\,2}(z)}{nhf(z)} \int_{-\infty}^{\infty} K^2(t)\mathrm{d}t + o(1).
\end{aligned}$$

From here, the proof follows since $\widehat{\Omega}_{uu}^*(z)\widehat{\Omega}_{vv}^*(z) = \Omega_{uu}^*(z)\Omega_{vv}^*(z) + o_P(1)$, $\widehat{\Omega}_{uv}^2(z) = \Omega_{uv}^{*\,2}(z) + o_P(1)$ and $\widehat{f}(z) = f(z) + o_P(1)$.

**Lemma 12.** *Assume that the conditions of Theorem 6 are satisfied. Furthermore, suppose that the bandwidth parameter $h \asymp n^{-1/4}$. We have for all $z \in \{z^i : i \in [n]\}$ that*
$$\left\|n^{3/8}\left(\widehat{T}(z) - \mathrm{vec}(\Omega^*(z)) + (\Omega^*(z) \otimes \Omega^*(z))\,\mathrm{vec}\left(\widehat{\Sigma}(z) - \Sigma^*(z)\right)\right)\right\|_\infty = o_P(1).$$

*Proof.* First by re-arranging the terms, we have

$$\widehat{T}(z) - \text{vec}(\Omega^*(z)) + (\Omega^*(z) \otimes \Omega^*(z))\, \text{vec}\left(\widehat{\Sigma}(z) - \Sigma^*(z)\right)$$
$$= \text{vec}\left(\widehat{\Omega}(z) - \Omega^*(z)\right) - \left(\widehat{\Omega}(z) \otimes \widehat{\Omega}(z)\right) \text{vec}\left(\Sigma^*(z) - \widehat{\Omega}(z)^{-1}\right) \quad (26)$$
$$+ \left(\Omega^*(z) \otimes \Omega^*(z) - \widehat{\Omega}(z) \otimes \widehat{\Omega}(z)\right) \text{vec}\left(\widehat{\Sigma}(z) - \Sigma^*(z)\right)$$

Let $\Delta(z) = \widehat{\Omega}(z) - \Omega^*(z)$, the first term of right hand side in (26) can be bounded by

$$\left\| \text{vec}\left(\widehat{\Omega}(z) - \Omega^*(z)\right) - \left(\widehat{\Omega}(z) \otimes \widehat{\Omega}(z)\right) \text{vec}\left(\Sigma^*(z) - \widehat{\Omega}(z)^{-1}\right) \right\|_\infty$$
$$= \left\| \widehat{\Omega}(z) - \Omega^*(z) - \widehat{\Omega}(z)\left(\Sigma^*(z) - \widehat{\Omega}(z)^{-1}\right)\widehat{\Omega}(z) \right\|_\infty$$
$$= \left\| \Delta(z)\Omega^*(z)^{-1}\Delta(z) \right\|_\infty \quad (27)$$
$$\leq \left\| \Delta(z) \right\|_\infty \left\| \Omega^*(z)^{-1}\Delta(z) \right\|_{\infty,\infty}$$
$$\leq \left\| \Delta(z) \right\|_\infty \left\| \Omega^*(z)^{-1} \right\|_{\infty,\infty} \left\| \Delta(z) \right\|_{\infty,\infty}$$
$$\leq d \left\| \Delta(z) \right\|_\infty^2 \left\| \Omega^*(z)^{-1} \right\|_{\infty,\infty}$$

and the second term of right hand side in (26) can be bounded by

$$\left\| \Omega^*(z) \otimes \Omega^*(z) - \left(\widehat{\Omega}(z) \otimes \widehat{\Omega}(z)\right) \text{vec}\left(\widehat{\Sigma}(z) - \Sigma^*(z)\right) \right\|_\infty$$
$$\leq \left\| \Omega^*(z) \otimes \Omega^*(z) - \widehat{\Omega}(z) \otimes \widehat{\Omega}(z) \right\|_{\infty,\infty} \left\| \widehat{\Sigma}(z) - \Sigma^*(z) \right\|_\infty$$
$$= \left\| \Omega^*(z) \otimes \Delta(z) + \Delta(z) \otimes \Omega^*(z) + \Delta(z) \otimes \Delta(z) \right\|_{\infty,\infty} \left\| \widehat{\Sigma}(z) - \Sigma^*(z) \right\|_\infty$$
$$= \left\| \Omega^*(z) \otimes \Delta(z) + \Delta(z) \otimes \Omega^*(z) + \Delta(z) \otimes \Delta(z) \right\|_{\infty,\infty} \left\| \widehat{\Sigma}(z) - \Sigma^*(z) \right\|_\infty \quad (28)$$
$$\leq \left(2\left\|\Omega^*(z) \otimes \Delta(z)\right\|_{\infty,\infty} + \left\|\Delta(z) \otimes \Delta(z)\right\|_{\infty,\infty}\right) \left\|\widehat{\Sigma}(z) - \Sigma^*(z)\right\|_\infty$$
$$\leq \left(2\left\|\Omega^*(z)\right\|_{\infty,\infty}\left\|\Delta(z)\right\|_{\infty,\infty} + \left\|\Delta(z)\right\|_{\infty,\infty}^2\right) \left\|\widehat{\Sigma}(z) - \Sigma^*(z)\right\|_\infty$$
$$\leq \left(2d\left\|\Delta(z)\right\|_\infty\left\|\Omega^*(z)\right\|_{\infty,\infty} + d^2\left\|\Delta(z)\right\|_\infty^2\right) \left\|\widehat{\Sigma}(z) - \Sigma^*(z)\right\|_\infty$$

Combining (27) and (28), together with $\|\Delta(z)\|_\infty = \mathcal{O}(n^{-3/8}\sqrt{\log p})$ and $\|\widehat{\Sigma}(z) - \Sigma^*(z)\|_\infty = \mathcal{O}(n^{-3/8}\sqrt{\log p})$ we obtain the conclusion since $\|\Omega^*(z)\|_{\infty,\infty}$ and $\|\Omega^*(z)^{-1}\|_{\infty,\infty}$ are upper bounded. □

## A.4 Proof of Lemma 1

(a)

$$\|\mathcal{H}^n(A(x))\|_{\infty,\infty} = \max_u \sum_v \sqrt{\frac{\sum_{i\in[n]} A_{uv}^2(x^i)}{n}} \leq \max_{i\in[n]} \max_u \sum_v A_{uv}(x^i) = \max_{i\in[n]} \|A(x^i)\|_{\infty,\infty}.$$

(b) $\forall u, v$, we have

$$\mathcal{H}_{uv}^n(A(x)) = \sqrt{\frac{\sum_{i\in[n]} A_{uv}^2(x^i)}{n}} \leq \max_{i\in[n]} \max_{u,v} A_{uv}(x^i) = \max_{i\in[n]} \|A(x^i)\|_{\infty}.$$

thus $\|\mathcal{H}^n(A(x))\|_{\infty} \leq \max_{i\in[n]} \|A(x^i)\|_{\infty}$.

(c)

$$\|\mathcal{H}^n(A(x)B(x))\|_{\infty,\infty} = \max_u \sum_v \sqrt{\frac{\sum_{i\in[n]}(\sum_k A_{uk}(x^i)B_{kv}(x^i))^2}{n}}$$

$$\leq \max_u \sum_v \sqrt{\frac{\sum_{i\in[n]}(\sum_k A_{uk}^2(x^i))(\sum_k B_{kv}^2(x^i))}{n}}$$

$$\leq \max_u \sum_v \sqrt{\frac{\sum_{i\in[n]}(\sum_k A_{uk}^2(x^i)) \max_k \sum_{i\in[n]} \sum_v B_{kv}^2(x^i)}{n^2}}$$

$$= \max_u \sqrt{\frac{\sum_{i\in[n]} \sum_k A_{uk}^2(x^i)}{n}} \max_k \sqrt{\frac{\sum_{i\in[n]} \sum_v B_{kv}^2(x^i)}{n}}$$

$$= \|\mathcal{H}^n(A(x))\|_{\infty,\infty} \|\mathcal{H}^n(B(x))\|_{\infty,\infty}.$$

(d) $\forall u, v$, we have

$$\mathcal{H}_{uv}^n(A(x)B(x)) = \sqrt{\frac{\sum_{i\in[n]}(\sum_k A_{uk}(x^i)B_{kv}(x^i))^2}{n}}$$

$$\leq \sum_k \sqrt{\frac{\sum_{i\in[n]} A_{uk}^2(x^i)}{n}} \sqrt{\frac{\sum_{i\in[n]} B_{kv}^2(x^i)}{n}}$$

$$\leq \max_k \sqrt{\frac{\sum_{i\in[n]} A_{uk}^2(x^i)}{n}} \sum_k \sqrt{\frac{\sum_{i\in[n]} B_{kv}^2(x^i)}{n}}$$

$$\leq \max_k \max_u \sqrt{\frac{\sum_{i\in[n]} A_{uk}^2(x^i)}{n}} \max_v \sum_k \sqrt{\frac{\sum_{i\in[n]} B_{vk}^2(x^i)}{n}}$$

$$= \|\mathcal{H}^n(A(x))\|_{\infty} \|\mathcal{H}^n(B(x))\|_{\infty,\infty}.$$

Thus $\|\mathcal{H}^n(A(x)B(x))\|_\infty \leq \|\mathcal{H}^n(A(x))\|_\infty \|\mathcal{H}^n(B(x))\|_{\infty,\infty}$.

(e) $\forall u, v$, we have

$$\begin{aligned}
\mathcal{H}^n_{uv}(A(x) + B(x)) &= \sqrt{\frac{\sum_{i \in [n]} (A_{uv}(x^i) + B_{uv}(x^i))^2}{n}} \\
&\leq \sqrt{\frac{\sum_{i \in [n]} A_{uv}(x^i)^2}{n}} + \sqrt{\frac{\sum_{i \in [n]} B_{uv}(x^i)^2}{n}} \\
&= \mathcal{H}^n_{uv}(A(x)) + \mathcal{H}_{uv}(B(x))
\end{aligned}$$

Thus $\|\mathcal{H}^n(A(x) + B(x))\|_\infty \leq \|\mathcal{H}^n(A(x)) + \mathcal{H}^n(B(x))\|_\infty$.

# References

F. Bach, R. Jenatton, J. Mairal, and G. Obozinski. Optimization with sparsity-inducing penalties. *Found. Trends Mach. Learn.*, 4(1):1–106, 2011.

O. Banerjee, L. El Ghaoui, and A. d'Aspremont. Model selection through sparse maximum likelihood estimation. *J. Mach. Learn. Res.*, 9(3):485–516, 2008.

A.-L. Barabási and R. Albert. Emergence of scaling in random networks. *Science*, 286(5439): 509–512, 1999.

S. P. Boyd, N. Parikh, E. Chu, B. Peleato, and J. Eckstein. Distributed optimization and statistical learning via the alternating direction method of multipliers. *Found. Trends Mach. Learn.*, 3(1):1–122, 2011.

T. T. Cai, W. Liu, and X. Luo. A constrained $\ell_1$ minimization approach to sparse precision matrix estimation. *J. Am. Stat. Assoc.*, 106(494):594–607, 2011.

T. T. Cai, W. Liu, and H. H. Zhou. Estimating sparse precision matrix: Optimal rates of convergence and adaptive estimation. *arXiv preprint arXiv:1212.2882*, 2012.

T. T. Cai, H. Li, W. Liu, and J. Xie. Covariate-adjusted precision matrix estimation with an application in genetical genomics. *Biometrika*, 100(1):139–156, 2013.

M. Chen, Z. Ren, H. Zhao, and H. H. Zhou. Asymptotically normal and efficient estimation of covariate-adjusted gaussian graphical model. *arXiv preprint arXiv:1309.5923*, 2013a.

X. Chen, M. Xu, and W. B. Wu. Covariance and precision matrix estimation for high-dimensional time series. *Ann. Stat.*, 41(6):2994–3021, 2013b.

J. Chiquet, Y. Grandvalet, and C. Ambroise. Inferring multiple graphical structures. *Stat. Comput.*, 21(4):537–553, 2011.

H. Chun, M. Chen, B. Li, and H. Zhao. Joint conditional Gaussian graphical models with multiple sources of genomic data. *Frontiers in Genetics*, 4, 2013.

P. Danaher, P. Wang, and D. M. Witten. The joint graphical lasso for inverse covariance estimation across multiple classes. *J. R. Stat. Soc. B*, 76(2):373–397, 2014.

A. P. Dempster. Covariance selection. *Biometrics*, 28:157–175, 1972.

J. C. Duchi, S. Gould, and D. Koller. Projected subgradient methods for learning sparse gaussians. In *Proc. of UAI*, pages 145–152, 2008.

J. Fan and W. Zhang. Statistical methods with varying coefficient models. *Statistics and its Interface*, 1(1):179–195, 2008.

J. H. Friedman, T. J. Hastie, and R. J. Tibshirani. Sparse inverse covariance estimation with the graphical lasso. *Biostatistics*, 9(3):432–441, 2008.

J. Guo, E. Levina, G. Michailidis, and J. Zhu. Joint estimation of multiple graphical models. *Biometrika*, 98(1):1–15, 2011.

P. Hoff and X. Niu. A covariance regression model. *Stat. Sinica*, 22:729–753, 2012.

J. Honorio and D. Samaras. Multi-task learning of gaussian graphical models. In J. Fürnkranz and T. Joachims, editors, *Proc. of ICML*, pages 447–454, Haifa, Israel, 2010. Omnipress.

J. Jankova and S. A. van de Geer. Confidence intervals for high-dimensional inverse covariance estimation. *ArXiv e-prints, arXiv:1403.6752*, 2014.

S. Janson. Maximal spacings in several dimensions. *Ann. Probab.*, 15(1):274–280, 1987.

A. Javanmard and A. Montanari. Confidence intervals and hypothesis testing for high-dimensional regression. *arXiv preprint arXiv:1306.3171*, 2013.

K. Khare, S.-Y. Oh, and B. Rajaratnam. A convex pseudo-likelihood framework for high dimensional partial correlation estimation with convergence guarantees. *arXiv preprint arXiv:1307.5381*, 2013.

M. Kolar and E. P. Xing. Sparsistent estimation of time-varying discrete markov random fields. *ArXiv e-prints, arXiv:0907.2337*, 2009.

M. Kolar and E. P. Xing. On time varying undirected graphs. In *Proc. of AISTATS*, 2011.

M. Kolar and E. P. Xing. Estimating networks with jumps. *Electron. J. Stat.*, 6:2069–2106, 2012.

M. Kolar, A. P. Parikh, and E. P. Xing. On sparse nonparametric conditional covariance selection. In J. Fürnkranz and T. Joachims, editors, *Proc. 27th Int. Conf. Mach. Learn.*, Haifa, Israel, 2010a.

M. Kolar, L. Song, A. Ahmed, and E. P. Xing. Estimating Time-varying networks. *Ann. Appl. Stat.*, 4(1):94–123, 2010b.

C. Lam and J. Fan. Sparsistency and rates of convergence in large covariance matrix estimation. *Ann. Stat.*, 37:4254–4278, 2009.

B. Li, H. Chun, and H. Zhao. Sparse estimation of conditional graphical models with application to gene networks. *J. Am. Stat. Assoc.*, 107(497):152–167, 2012.

H. Li and J. Gui. Gradient directed regularization for sparse gaussian concentration graphs, with applications to inference of genetic networks. *Biostatistics*, 7(2):302–317, 2006.

Q. Li and J. S. Racine. *Nonparametric Econometrics: Theory and Practice*. Princeton University Press, 2006.

J. Liu, R. Li, and R. Wu. Feature selection for varying coefficient models with ultrahigh dimensional covariates. *Technical Report*, 2013.

W. Liu. Gaussian graphical model estimation with false discovery rate control. *arXiv preprint arXiv:1306.0976*, 2013.

R. Mazumder and T. J. Hastie. Exact covariance thresholding into connected components for large-scale graphical lasso. *J. Mach. Learn. Res.*, 13:781–794, 2012.

N. Meinshausen and P. Bühlmann. High dimensional graphs and variable selection with the lasso. *Ann. Stat.*, 34(3):1436–1462, 2006.

K. Mohan, P. London, M. Fazel, D. M. Witten, and S.-I. Lee. Node-based learning of multiple gaussian graphical models. *J. Mach. Learn. Res.*, 15:445–488, 2014.

Y. Nesterov. A method of solving a convex programming problem with convergence rate $\mathcal{O}(1/k^2)$. In *Soviet Mathematics Doklady*, volume 27, pages 372–376, 1983.

F. Orabona, A. Argyriou, and N. Srebro. PRISMA: PRoximal iterative SMoothing algorithm. *ArXiv e-prints, arXiv:1206.2372*, 2012.

A. Pagan and A. Ullah. *Nonparametric Econometrics (Themes in Modern Econometrics)*. Cambridge University Press, 1999.

B. U. Park, E. Mammen, Y. K. Lee, and E. R. Lee. Varying coefficient regression models: A review and new developments. *International Statistical Review*, pages n/a–n/a, 2013.

J. Peng, P. Wang, N. Zhou, and J. Zhu. Partial correlation estimation by joint sparse regression models. *J. Am. Stat. Assoc.*, 104(486):735–746, 2009.

M. Pourahmadi. *High-Dimensional Covariance Estimation: With High-Dimensional Data (Wiley Series in Probability and Statistics)*. Wiley, 2013.

P. Ravikumar, M. J. Wainwright, G. Raskutti, and B. Yu. High-dimensional covariance estimation by minimizing $\ell_1$-penalized log-determinant divergence. *Electron. J. Stat.*, 5: 935–980, 2011.

Z. Ren, T. Sun, C.-H. Zhang, and H. H. Zhou. Asymptotic normality and optimalities in estimation of large gaussian graphical model. *arXiv preprint arXiv:1309.6024*, 2013.

G. V. Rocha, P. Zhao, and B. Yu. A path following algorithm for sparse pseudo-likelihood inverse covariance estimation (splice). *arXiv preprint arXiv:0807.3734*, 2008.

A. J. Rothman, P. J. Bickel, E. Levina, and J. Zhu. Sparse permutation invariant covariance estimation. *Electron. J. Stat.*, 2:494–515, 2008.

B. W. Silverman. Weak and strong uniform consistency of the kernel estimate of a density and its derivatives. *Ann. Stat.*, 6(1):177–184, 1978.

T. Sun and C.-H. Zhang. Sparse matrix inversion with scaled lasso. 2012.

Y. Tang, H. J. Wang, Z. Zhu, and X. Song. A unified variable selection approach for varying coefficient models. *Stat. Sinica*, 22(2), 2012.

S. A. van de Geer, P. Bühlmann, Y. Ritov, and R. Dezeure. On asymptotically optimal confidence regions and tests for high-dimensional models. *arXiv preprint arXiv:1303.0518*, 2013.

G. Varoquaux, A. Gramfort, J.-B. Poline, and B. Thirion. Brain covariance selection: Better individual functional connectivity models using population prior. In J. D. Lafferty, C. K. I. Williams, J. Shawe-Taylor, R. S. Zemel, and A. Culotta, editors, *Proc. of NIPS*, pages 2334–2342, 2010.

R. Vershynin. Introduction to the non-asymptotic analysis of random matrices. In Y. C. Eldar and G. Kutyniok, editors, *Compressed Sensing: Theory and Applications*. Cambridge University Press, 2012.

M. J. Wainwright. Sharp thresholds for high-dimensional and noisy sparsity recovery using $\ell_1$-constrained quadratic programming (lasso). *IEEE Trans. Inf. Theory*, 55(5):2183–2202, 2009.

L. Wang, H. Li, and J. Z. Huang. Variable selection in nonparametric varying-coefficient models for analysis of repeated measurements. *J. Am. Stat. Assoc.*, 103(484):1556–1569, 2008.

D. M. Witten, J. H. Friedman, and N. Simon. New insights and faster computations for the graphical lasso. *J. Comput. Graph. Stat.*, 20(4):892–900, 2011.

J. Yin, Z. Geng, R. Li, and H. Wang. Nonparametric covariance model. *Statistica Sinica*, 20: 469–479, 2010.

J. Yin and H. Li. A sparse conditional Gaussian graphical model for analysis of genetical genomics data. *Ann. Appl. Stat.*, 5(4):2630–2650, 2011.

J. Yin and H. Li. Adjusting for high-dimensional covariates in sparse precision matrix estimation by 1-penalization. *J. Multivar. Anal.*, 116:365–381, 2013.

M. Yuan. High dimensional inverse covariance matrix estimation via linear programming. *J. Mach. Learn. Res.*, 11:2261–2286, 2010.

M. Yuan and Y. Lin. Model selection and estimation in regression with grouped variables. *J. R. Stat. Soc. B*, 68:49–67, 2006.

M. Yuan and Y. Lin. Model selection and estimation in the gaussian graphical model. *Biometrika*, 94(1):19–35, 2007.

X. Yuan. Alternating direction method for covariance selection models. *J. Sci. Comp.*, 51 (2):261–273, 2012.

C.-H. Zhang and S. S. Zhang. Confidence intervals for low-dimensional parameters in high-dimensional linear models. 2011.

W. Zhang and S.-Y. Lee. Variable bandwidth selection in varying-coefficient models. *J. Multivar. Anal.*, 74(1):116–134, 2000.

P. Zhao and B. Yu. On model selection consistency of lasso. *J. Mach. Learn. Res.*, 7: 2541–2563, 2006.

T. Zhao, H. Liu, K. E. Roeder, J. D. Lafferty, and L. A. Wasserman. The huge package for high-dimensional undirected graph estimation in R. *J. Mach. Learn. Res.*, 13:1059–1062, 2012.

S. Zhou, J. D. Lafferty, and L. A. Wasserman. Time varying undirected graphs. *Mach. Learn.*, 80(2-3):295–319, 2010.

Y. Zhu, X. Shen, and W. Pan. Structural pursuit over multiple undirected graphs. *Technical Report*, 2013.